%% file: main.tex
\documentclass[10pt,twocolumn,letterpaper]{article}

\usepackage{cvpr}    


\definecolor{cvprblue}{rgb}{0.21,0.49,0.74}
\usepackage[pagebackref,breaklinks,colorlinks,allcolors=cvprblue]{hyperref}
\usepackage{xcolor}
\usepackage{listings}   
\usepackage{multirow}
\usepackage{algorithm}
\usepackage{algorithmic}
\usepackage{wrapfig}
\usepackage{hyperref}
\usepackage{url}
\usepackage{adjustbox}
\usepackage{graphicx}
\usepackage{amssymb}
\usepackage{subcaption}

\usepackage{array}      
\usepackage{booktabs}   
\usepackage{caption}   
\usepackage{float}
\usepackage{dblfloatfix}
\usepackage{xcolor}
\usepackage{subcaption}
\usepackage{tabularx}
\usepackage{makecell}
\usepackage[table]{xcolor}
\usepackage{pifont}
\usepackage[table,dvipsnames,svgnames]{xcolor}

\title{Perceptual-Evidence Anchored Reinforced Learning for Multimodal Reasoning}

\author{Chi Zhang\textsuperscript{1}, Haibo Qiu\textsuperscript{2}\footnotemark, Qiming Zhang\textsuperscript{3}, 
Yufei Xu\textsuperscript{3},
Zhixiong Zeng\textsuperscript{2},
\\
Siqi Yang\textsuperscript{2},Peng Shi\textsuperscript{2} 
Lin Ma\textsuperscript{2$\dagger$}, Jing Zhang\textsuperscript{1$\dagger$} \\
\textsuperscript{1}School of Computer Science, Wuhan University\\
\textsuperscript{2}Meituan Inc\\
\textsuperscript{3}The University of Sydney \\ 
\texttt{forest.linma@gmail.com, jingzhang.cv@gmail.com} \\
\textbf{Project: \url{https://github.com/MiliLab/PEARL}}
}

\begin{document}

\twocolumn[{%
\renewcommand\twocolumn[1][]{#1}%
\maketitle
\begin{center}
\begin{minipage}{0.63\linewidth}
  \centering
  \includegraphics[width=\linewidth]{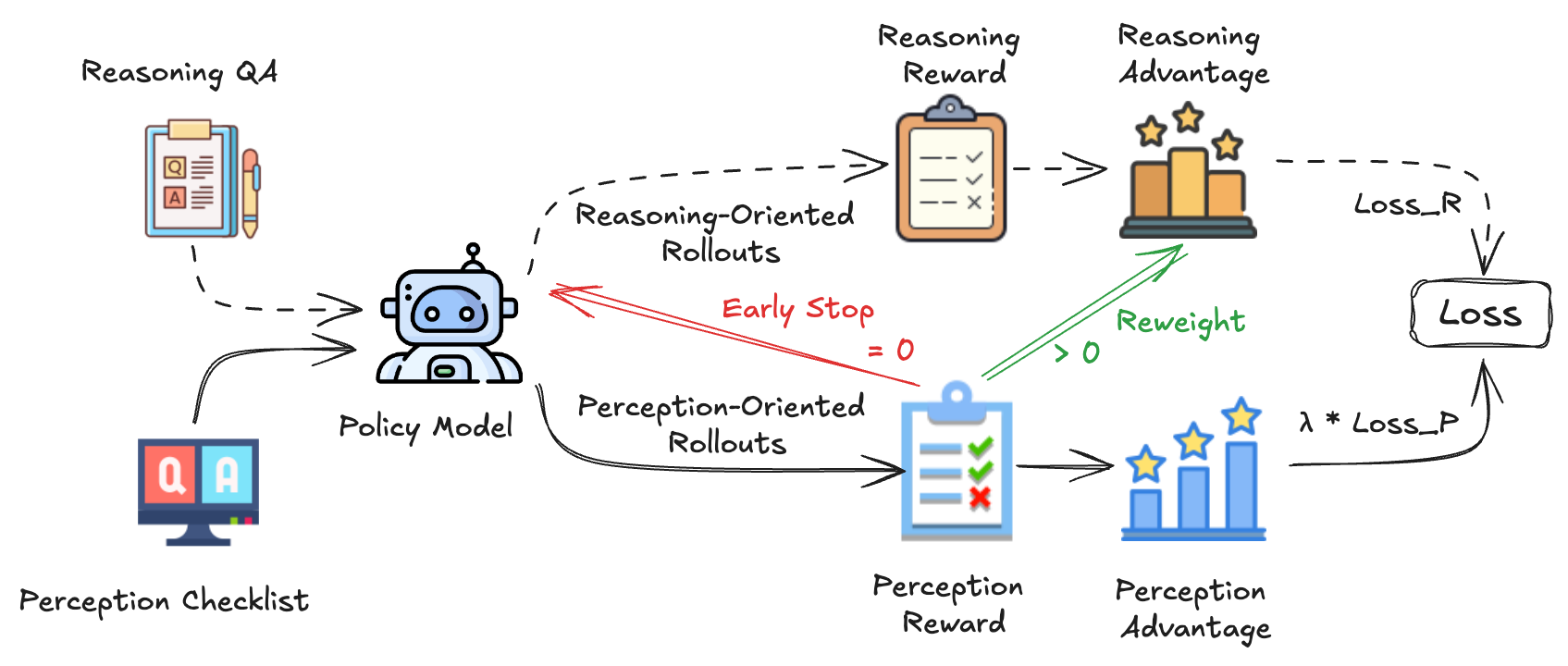}
\end{minipage}%
\hfill
\begin{minipage}{0.35\linewidth}
  \centering
  \raisebox{2mm}{\includegraphics[width=\linewidth]{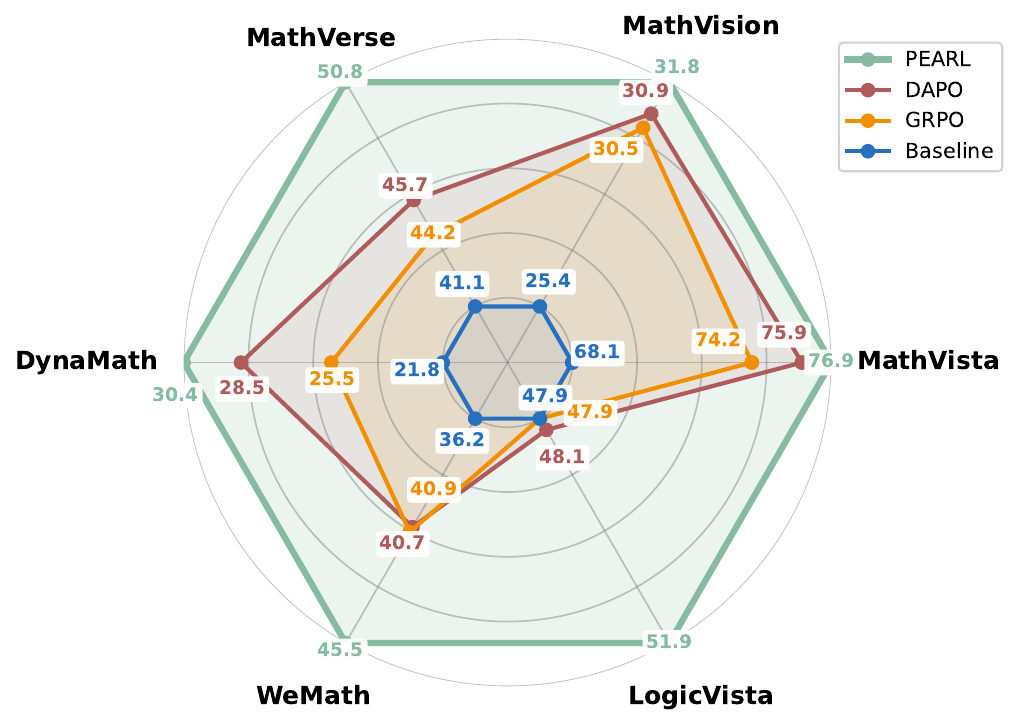}}
\end{minipage}
\captionof{figure}{We present PEARL, a perception–reasoning synergistic RL paradigm that strengthens VLMs' reasoning by explicitly anchoring in perceptual evidence, and demonstrates outstanding performance across multiple multimodal reasoning benchmarks.
\vspace{1em}
}
\label{fig:teaser}
\end{center}
}]
{
\renewcommand\thefootnote{}%
\footnotetext{* Project Leader. $\dagger$ Corresponding author.}%
\addtocounter{footnote}{-1}%
}
\input{sec/0_abstract}    
\input{sec/1_intro}
\input{sec/2_relatedw}
\input{sec/3_method}
\input{sec/4_exp}
\input{sec/5_conclusion}
\input{sec/X_suppl}

\clearpage
{
    \small
    \bibliographystyle{ieeenat_fullname}
    \bibliography{main}
}

\end{document}

%% file: sec/0_abstract.tex
\begin{abstract}
Reinforcement Learning with Verifiable Rewards (RLVR) has significantly advanced the reasoning capabilities of Large Language Models (LLMs) and is now being applied to Vision-Language Models (VLMs). However, vanilla RLVR for VLMs verifies only the final textual output, critically neglecting the foundational step of visual perception. This oversight leads to visual hallucinations and reward hacking, as reasoning built upon flawed perception is inherently unreliable. To address this, we propose \textbf{PEARL} (\textbf{P}erceptual-\textbf{E}vidence \textbf{A}nchored \textbf{R}einforced \textbf{L}earning), a dual-branch, perception-reasoning synergistic that strengthens multimodal reasoning by explicitly anchoring it to verified visual evidence. For each reasoning-oriented QA instance, PEARL first derive a \textit{perception checklist}---a set of perception-oriented sub-questions with verifiable answers that probe the model's understanding of key visual evidence. During training, auxiliary rollouts on this checklist yield a perceptual reward that both directly reinforces the model's perception ability and acts as a fidelity gate for reasoning. If the model passes the perception check, its policy update is biased towards evidence-anchored reasoning. Otherwise, the process is halted to prevent reasoning from flawed premises. PEARL can be seamlessly integrated with popular RL methods like GRPO and DAPO. Comprehensive experiments show PEARL achieves substantial gains on multimodal reasoning benchmarks, e.g., a +9.7\% improvement over the baseline and +6.6\% over GRPO on MathVerse.
\end{abstract}

%% file: sec/1_intro.tex
\section{Introduction}
\label{sec:intro}
The success of Reinforcement Learning with Verifiable Rewards (RLVR) in enhancing the reasoning ability of Large Language Models (LLMs) has spurred its application to Vision-Language Models (VLMs) for complex multimodal tasks. While this line of work shows promise on multimodal reasoning benchmarks~\cite{huang2025vision,chen2025vinci,zhang2025r1,tan2025reason,chen2025sft,wang2025vl}, it harbors a fundamental flaw: RLVR is applied in an outcome-only fashion, verifying the final textual answer while critically ignoring the fidelity of the upstream visual perception.~\cite{zhang2025r1,li2025self,huang2025boosting,wang2025perception}. This oversight leads to models that exploit spurious correlations and exhibit reward hacking, as their reasoning is not anchored to verified visual evidence. The resulting entanglement of flawed perception and downstream reasoning produces visually ungrounded outputs and poses a significant barrier to further advance multimodal reasoning~\cite{liu2025more,yao2025reasoning}.

\begin{figure*}[ht]
    \centering
    \includegraphics[width=1\linewidth]{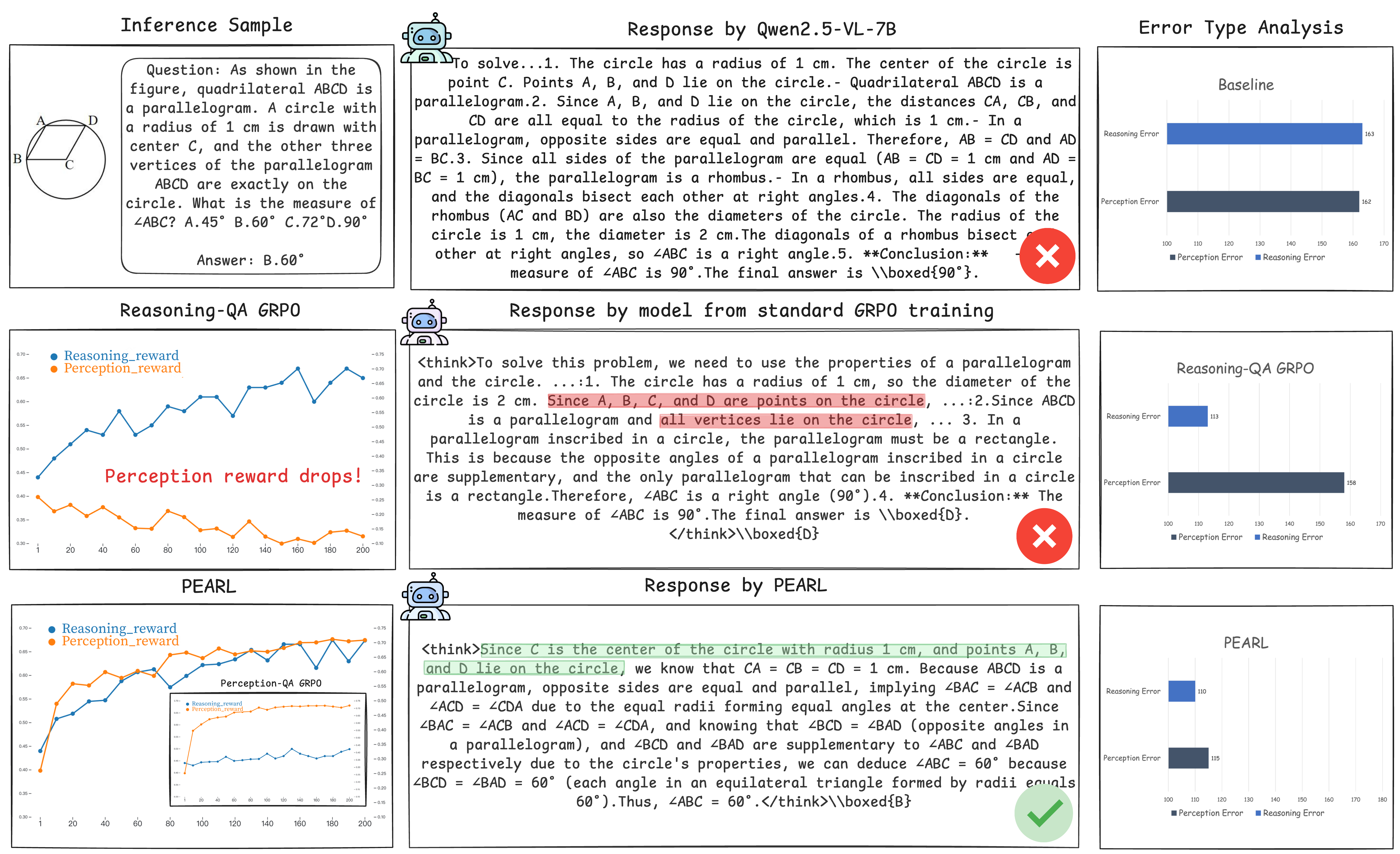}
    \caption{Comparison of training dynamics and failure modes. Standard outcome-based GRPO reduces reasoning errors but fails to address fundamental perception errors, leading to spurious reasoning chains based on flawed visual understanding. In contrast, our proposed PEARL achieves simultaneous improvements in both perception and reasoning, leading to a marked reduction in both error types and enabling more reliable problem-solving.}
    \label{fig:survey}
\end{figure*}

To empirically validate this claim, we conduct a preliminary diagnostic analysis. We finetune a leading VLM~\cite{qwen2_5_vl} using a representative outcome-based RLVR method, GRPO~\cite{shao2024deepseekmath,guo2025deepseek}, and compare its failure modes against the original model on WeMath~\cite{qiao2025we}. We introduce a taxonomy to disentangle failures into (i) perception errors, where the model misinterprets visual facts (e.g., object attributes, diagram elements), and (ii) reasoning errors, which involve flawed logic or incorrect calculations.

The results, summarized in Fig.~\ref{fig:survey}, are telling. Standard GRPO training leads to a substantial decrease in reasoning errors, demonstrating its effectiveness in refining step-by-step problem-solving abilities. However, this success is deceptive. In stark contrast, the frequency of perception-related errors remains almost unchanged. This divergence creates spurious reasoning chains: the model learns to generate logically plausible steps based on fundamentally flawed visual understanding, simply because the final outcome is correct. This evidence strongly supports our hypothesis: existing RLVR methods optimize for correct reasoning procedures but are blind to the perceptual errors they are built upon, creating a critical bottleneck that imposes a hard ceiling on both performance and reliability.

In response to this challenge, we propose \textbf{PEARL} (\textbf{P}erceptual-\textbf{E}vidence \textbf{A}nchored \textbf{R}einforced \textbf{L}earning), a perception-reasoning synergistic RL paradigm designed to strengthen multimodal reasoning through explicitly cultivating perceptual competence and promoting reasoning that is firmly anchored in perceptual evidence. Instead of treating the VLM as a black box that produces a final answer, PEARL begins by asking: ``Did the model even see the image correctly before reasoning?'' To answer this, for any given reasoning problem, PEARL first derives a perception checklist—a series of simple, factual questions about the key visual evidence required for reasoning.

This checklist serves a dual purpose within our reinforcement learning scheme. First, by generating auxiliary rollouts on these sub-questions, we obtain a direct perceptual reward that explicitly reinforces the model's ability to accurately extract visual information. Second, and more critically, the outcome of this check acts as a fidelity gate. Only if the model demonstrates sufficient perceptual accuracy by ``passing'' the checklist is the policy update for the main reasoning task permitted to proceed. Conversely, if the model fails the perception check, the reasoning update is halted. This mechanism is crucial: it prevents the model from reinforcing spurious reasoning chains, directly combating reward hacking. In essence, PEARL compels the model to learn how to reason correctly only when it has proven it can see correctly.

The direct impact of this design is demonstrated back in our diagnostic analysis (Fig.~\ref{fig:survey}). Unlike GRPO, PEARL achieves concurrent and substantial reductions in both perception and reasoning errors, successfully breaking the performance ceiling imposed by faulty perception. Furthermore, our results show that even training with the perceptual reward alone yields notable improvements in reasoning (validated in Tab.~\ref{tab:generalization}), unequivocally confirming that perceptual accuracy is a critical and foundational prerequisite for high-level reasoning. By forging this robust and verifiable link between seeing and thinking, PEARL paves the way for more reliable and capable multimodal reasoning models. In summary, our contributions are threefold:
\begin{itemize}[leftmargin=1.5em]
    \item We identify and empirically demonstrate a critical flaw in existing RLVR methods for VLMs: their blindness to upstream perception errors, which imposes a hard ceiling on model performance by reinforcing reasoning based on flawed visual perception.
    \item We propose PEARL, a novel RL framework that forges a verifiable link between perception and reasoning. PEARL uses a perception checklist to provide direct perceptual rewards and act as a fidelity gate, preventing policy updates for reasoning tasks when the underlying perception is incorrect.
    \item We demonstrate the significant effectiveness and generality of our approach. PEARL achieves substantial gains on challenging multimodal reasoning benchmarks,  including a +9.7\% absolute improvement on MathVerse, validating that anchoring reasoning to verified perception is a crucial step forward.
\end{itemize}

%% file: sec/2_relatedw.tex
\section{Related Work}
\label{sec:Related Work}
RLVR has significantly advanced the reasoning abilities of LLMs. More recently, the success of methods like GRPO~\cite{guo2025deepseek,shao2024deepseekmath} has spurred a wave of investigation into VLMs, leading to notable achievements~\cite{meng2025mm,wang2025vl,deng2025openvlthinker,shen2025vlm}. Alongside this progress, a growing consensus recognizes that many reasoning failures in VLMs stem from a critical gap: insufficient perceptual grounding and image understanding~\cite{zhang2025r1,wang2025perception,li2025self,thawakar2025llamav,xiao2025advancing,xia2025visionary,zhang2025thyme,chen2025mint,yang2025look,zhang2025deepsketcher,zheng2025deepeyes}. One prominent line of work attempts to address this by allowing models to interact directly with visual input, for example, by selecting regions, drawing, or highlighting key visual cues~\cite{zhang2025thyme,chen2025mint,yang2025look,zhang2025deepsketcher,zheng2025deepeyes}. These interactive mechanisms encourage models to attend more carefully to task-relevant visual content and to reason about finer details. However, such approaches typically rely on carefully curated training data to these interactive skills. Consequently, their performance is sensitive to data quality and scale, which limits their general applicability.

Another line of research modifies reinforcement learning to implicitly align perception with reasoning~\cite{wang2025perception,li2025self}. Typical strategies decouple perception and reasoning via an intermediate image-description stage, or probe models under altered visual states (e.g., masked input) to diagnose and penalize perception–reasoning discrepancies. These can be framed as condition-based alignment methods: they constrain or evaluate reasoning under varying visual conditions to ensure that the model's conclusions remain effectively coupled with the perceived input.

Seeking a more controllable learning signal, other approaches encourage models to generate structured image descriptions or captions prior to reasoning~\cite{zhang2025r1,thawakar2025llamav,xiao2025advancing,xia2025visionary}. This ``describe-then-reason'' paradigm aims to enable direct evaluation of visual understanding via explicit reward signals, often derived from external reward models or heuristic scoring functions. However, as previously discussed, such reward signals are frequently noisy and difficult to optimize, as quantifying the ``correctness'' of free-form text is inherently ambiguous. Furthermore, relying on additional reward models or LLM-based judges introduces significant computational and implementation overhead, posing a major barrier to large-scale deployment.

%% file: sec/3_method.tex
\section{Method}

\begin{figure*}[!t]
    \centering
    \includegraphics[width=1\linewidth]{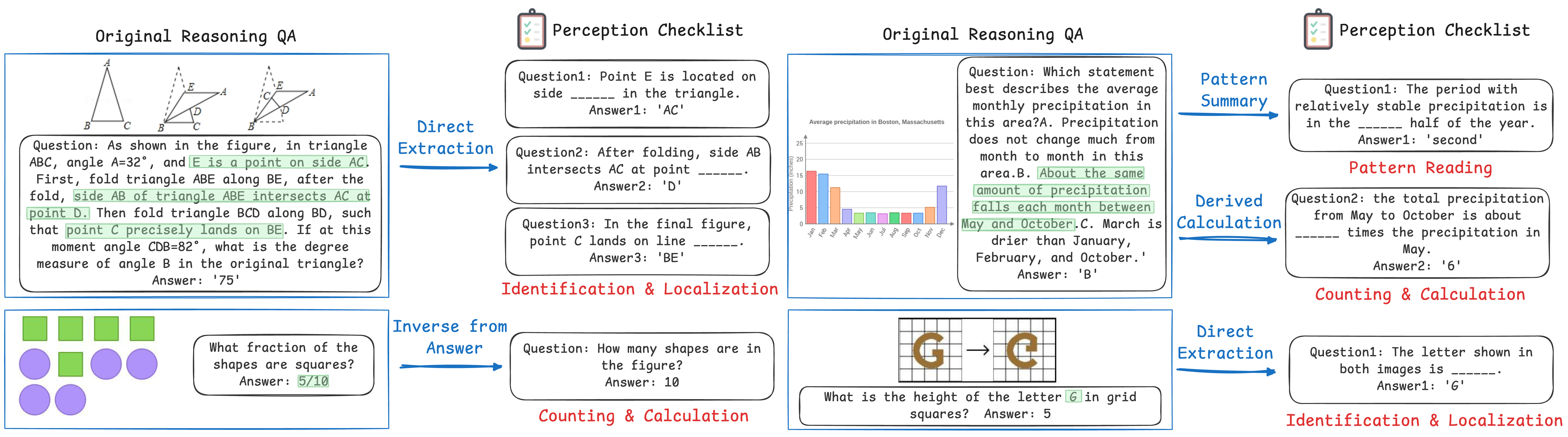}
    \caption{Representative sub-question/answer pairs produced under our guidelines from original image–QA instances. These task-aligned, image-grounded items provide high-quality, easily verifiable supervision and serve as perceptual scaffolds supporting reasoning.}
    \label{fig:pearl-cases}
\end{figure*}

As shown in Fig.~\ref{fig:teaser}, PEARL is a perception-reasoning synergistic framework that explicitly anchors the reasoning in perceptual evidence to achieve more reliable and capable multimodal reasoning. In the following, we first briefly revisit the formulation of GRPO and identify the flaw in only verifying the textual outcome; then we demonstrate how the perception and reasons rollouts performed and collaborated for policy optimization.  

\subsection{Prerequisites}
\label{sec:problem setting}
Our work is built upon the  Group Relative Policy Optimization (GRPO) under the RLVR paradigm~\cite{guo2025deepseek}. Given a multimodal query $Q = (I, q)$ (consisting of an image $I$ and a textual question $q$), the policy model $\pi_{\theta_{\text{old}}}$ samples a group of $G$ candidate outputs $\{o_i\}_{i=1}^G$. A rule-based verifier then assigns each response ${o_i}$ a scalar reward $r_i$ reflecting its correctness. The optimization goal is to maximize the following clipped objective:
{\small
\begin{equation}
\begin{aligned}
\mathcal{J}_{\text{GRPO}}(\theta)
&= \mathbb{E}_{Q \sim \mathcal{D},\,\{o_i\}_{i=1}^G 
\sim \pi_{\theta_{\text{old}}}(\cdot \mid Q)} \Bigg[ \\
&\qquad\hspace{-5em}
\frac{1}{G}\sum_{i=1}^{G}\frac{1}{|o_i|}\sum_{t=1}^{|o_i|}
\min\!\left(
    r_{i,t}(\theta)\hat{A}_{i,t}, \mathrm{clip}\!\left(r_{i,t}(\theta), 1-\epsilon, 1+\epsilon\right)\hat{A}_{i,t}
\right)
\\
&\qquad\qquad - \beta D_{\mathrm{KL}}\!\left(\pi_\theta \,\|\, \pi_{\text{ref}}\right)
\Bigg]
\end{aligned}
\label{eq:eq2}
\end{equation}
}
where $\beta$ controls the KL regularization strength, while the advantage and importance sampling ratio are defined as: 
{\small
\begin{equation}
    \hat{A}_i
    = \frac{r_i- \text{mean}(\{r_i\}_{i=1}^G)}{\text{std}(\{r_i\}_{i=1}^G)}, \ r_{i,t}(\theta) = \frac{\pi_\theta(o_{i,t} \mid q,o_{i,<t})}{\pi_{\theta_{\text{old}}}(o_{i,t} \mid q,o_{i,<t})}
\label{eq:eq1}
\end{equation}
}
It is clear that the reward is defined solely by the final textual outcome while neglecting perceptual evidence, leading to the risk of visual hallucinations and spurious reasoning.

\subsection{Perception Checklist Construction}
\label{sebsec:probes construction}
A common method for evaluating image perception involves instructing models to generate detailed captions or free-form descriptions, scored by external reward models or LLM judges~\cite{zhang2025r1,yao2024mulberry,xia2025visionary}. However, this approach faces practical limitations: description-based rewards are inherently noisy as scoring lengthy and unstructured text is highly sensitive to phrasing and redundancy, weakening the alignment between the reward signal and the visual understanding required for reasoning tasks. This ambiguity often results in perception reward hacking.

Instead, PEARL moves from dense, free-form descriptions to task-aligned perception checklist derived directly from corresponding QA pairs. We generate these perception probes by operating along two primary dimensions to ensure they are both semantically grounded and perceptually specific. 

The first dimension focuses on the content derivation. Given a multimodal reasoning instance, we develop a set of ``operation guidelines" to define how to discover ``question-worthy'' content embedded in the instance. These guidelines instruct the model to apply one of four derivation patterns: (1) Direct Extraction, which surfaces facts explicitly presented in the image or question; (2) Pattern Summary, which isolates salient regions or structural cues essential for solving the original problem; (3) Derived Calculation, which requires a one-step inference or computation based on visual patterns; and (4) Inverse from Answer, which treats the original answer as a constraint to recover latent counts or relations. 

The second dimension priorities the targeted perceptual skill, which defines what the resulting sub-question will probe, targeting a specific low-level skill such as identifying objects, reading trends, counting quantities, or recognizing geometric arrangements (see Fig. \ref{fig:pearl-cases} for examples).

This dual-dimensions design does more than just prompt multiple QA pairs; it is engineered to provide two key advantages. First, the generated sub-questions are inherently, logically tied to the original task, allowing them to serve as a ``perception checklist'' to verify the model's requisite understanding. Second, their answers are typically concise and high-quality verifiable (e.g., a number or label), enabling RLVR training.

\subsection{Perception and Reasoning Rollouts}
As illustrated in Fig.~\ref{fig:teaser}, we begin with a reasoning instance $(Q_r, A_r)$. Here, $Q_r = (I, q_r)$ is the multimodal query (as defined in Sec.~\ref{sec:problem setting}), consisting of the image $I$ and the primary reasoning question $q_r$. $A_r$ is the corresponding ground-truth reasoning answer. To each instance $Q_r$, we attach a perception checklist including the set of perception QA probes $\{(Q_p^i, A_p^i)\}_{i=1}^{K}$ constructed as described in Sec.~\ref{sebsec:probes construction}. At each training step, we have two distinct prompts: (i) a reasoning prompt $Q_r$ which instructs the VLM to perform step-by-step reasoning, and (ii) a compact perception prompt $\tilde{Q}_p$ that serializes all $K$ perception probes into a single structured sequence: $\tilde{Q}_p= \big[I, Q_p^1,\; \ldots,\; Q_p^{K}\big].$

\paragraph{Perception-Oriented Rollouts.}
The VLM is first prompted to rollout on $\tilde{Q}_p$ multiple times to directly answer those probes $\{Q_p^i\}_{i=1}^{K}$ without intermediate reasoning. G outputs $\{o_p^i\}_{i=1}^G$ are generated with $o_p^i= \big[\hat{A}_p^{i,1},\; \ldots,\; \hat{A}_p^{{i,K}}\big]$. Then we parse the per-probe prediction $\hat{A}_p^{i,j}$ and employ a rule-based verifier to 
assign probe-level rewards $\{R_p^{i,j}\}_{j=1}^{K}$. The reward for $o_p^i$ is calculated as: $R_p^i = \frac{1}{K}\sum_{j=1}^{K}R^{i, j}$. Finally, We average this across all $G$ outputs to compute $\bar{R}_p$, which measures the model's perception on the image $I$ and also acts as a fidelity gate for the following reasoning rollouts.
\paragraph{Reasoning-Oriented Rollouts.}
\label{sec:Reasoning-Oriented Rollouts}
As demonstrated in Fig.~\ref{fig:teaser}, by leveraging $\bar{R}_p$, we design an early-stop filtering strategy to prevent the model from learning spurious reasoning patterns based on flawed perception and also reduce training costs. Specifically, for a given instance, if the fidelity gate $\bar{R}_p = 0$, we consider the current model as lacking sufficient perceptual competence to support the associated reasoning task. In this case, PEARL does not perform the following reasoning rollouts and moves to next training step. Only when the fidelity gate $\bar{R}_p > 0$, which is regarding as ``perception-passes'' instance, we allow the reasoning rollouts continues (as described in Sec.~\ref{sec:problem setting}) to obtain the reasoning reward $R_r$.

This filtering mechanism effectively prioritizes updates on examples where the model has at least a minimal perceptual foundation. As training progresses and perceptual performance improves, more instances satisfy this criterion, yielding a natural progression from ``learning to see'' to handling harder reasoning cases without requiring an explicit hand-crafted curriculum.

\subsection{Synergistic Optimization}

\paragraph{Perception-Reweight Reasoning.}
To further anchor reasoning optimization in perceptual feedback, we introduce a perception-reweight mechanism that modulates the reasoning advantage. For each reasoning rollout, we treat the mean perceptual reward $\bar{R}_p$ as a soft reliability prior that measures whether the accompanying reasoning trace is anchored in faithful perceptual understanding rather than spurious cues. Concretely, we reshape the group-normalized reasoning advantage $\hat{A}_r$ as: 
\begin{equation}
\hat{A}_r\leftarrow \hat{A}_r \cdot \mathrm{min}(\bar{R}_p,\ 0.5),
\end{equation}
where the perceptual reward acts as a confidence signal that scales the reasoning gradient. This mechanism amplifies perception-anchored reasoning while suppressing updates that are inconsistent with the perceptual evidence. Under the mild assumption that visually ungrounded, reward-hacking behaviors tend to induce systematically low $\bar{R}_p$, this mechanism biases optimization toward policies that are both textually successful and perceptually consistent, thereby reducing the relative competitiveness of purely spurious solutions.

\paragraph{Soft Online Filtering.}
We further adopt an online filtering scheme inspired by~\cite{yu2025dapo} with a key modification. The original scheme retains only samples whose mean reasoning reward satisfies $\bar{R}_r \notin [0,1]$, to avoid degenerated cases (i.e., zero reward variance) that yield no policy gradient. 
In our setting, this strict criterion would undesirably discard instances where reasoning rewards $\bar{R}_r$ have saturated (all 0s or all 1s) but the perception signal remains informative. To prevent this loss of signal, we relax the criterion and retain a sample group whenever: 
\begin{equation}
\bar{R}_r \notin [0,1] \;\lor\; \bar{R}_p \notin [0,1].
\end{equation}
This ensures that samples with a non-trivial signal from either path contribute to learning, while still filtering out fully saturated cases that provide no useful gradient.

\paragraph{Dual-Objective Formulation.}
We finally extend the GRPO objective in Eq.~\ref{eq:eq2} to drive two coordinated optimization paths, enforcing both accurate perception and coherent reasoning. Let $\hat{A}_p$ be the group-normalized advantage 
from the perception rewards $\{R_p^i\}_{i=1}^G$. We denote by $\mathcal{J}_{\mathrm{GRPO}}(\theta; \hat{A})$ 
the GRPO objective parameterized by a given advantage $\hat{A}$. Our dual-objective formulation is defined as:
\begin{equation}
\mathcal{J}_{\text{dual}}(\theta)
= \mathcal{J}_{\mathrm{GRPO}}(\theta; \hat{A}_r)
+ \lambda \mathcal{J}_{\mathrm{GRPO}}(\theta; \hat{A}_p),
\label{eq:dual_objective}
\end{equation}
where $\lambda$ controls the relative contribution of the perception
path in the overall gradient update. 

%% file: sec/4_exp.tex
\begin{table*}[!t]
\centering
\caption{Performance comparison on multimodal reasoning benchmarks. The best results are highlighted in \textbf{bold}. We separately group methods that emphasize perception-aware optimization (e.g., Vision-SR1 and PEARL) below the line in each section.}
\resizebox{\linewidth}{!}{
\begin{tabular}{lccccccc}
\hline
\textbf{Model} & \textbf{MathVerse} & \textbf{Mathvision} & \textbf{MathVista} & \textbf{LogicVista} & \textbf{WeMath} & \textbf{DynaMath} & \textbf{Average} \\
\hline
\rowcolor{gray!10} \multicolumn{8}{c}{\text{Based on Qwen2.5-VL-3B}} \\

Base Model & 31.2 & 21.9 & 61.2 & 40.3& 22.9 & 13.2 &31.8\\
GRPO~\cite{shao2024deepseekmath} & 34.9 & 26.8 & 64.7 & 37.4 & 26.9 & 18.2 & 34.8\\
DAPO~\cite{yu2025dapo} & 36.0 & 27.1 & 65.4 & 41.2 & 32.1 & 18.2 & 36.6\\
VLAA-Thinker~\cite{chen2025sft} & 36.4 & 24.4 & 61.0 & 38.5& 33.8 & 18.2 &35.4\\
\hline
Visionary-R1~\cite{xia2025visionary} & 32.3& 24.3 & 64.8& 38.7 & 28.6 & 13.0 & 33.6 \\
$\text{PAPO}_G$~\cite{wang2025perception} & 36.1 &  26.4 & 64.6&40.4 & 29.8 & 18.9 & 36.0\\
$\text{PAPO}_D$~\cite{wang2025perception} & 40.1 & 27.0 & 67.0 & 45.1& 34.9 & 21.1 & 39.2\\
Vision-SR1~\cite{li2025self} & 37.4 & 26.4 & 65.4 & 43.8 & 32.2 & 18.2 & 37.2\\
PEARL & \textbf{40.5} & \textbf{27.8} & \textbf{67.1} & \textbf{45.6} & \textbf{36.3} & \textbf{21.4} & \textbf{39.8}\\
\hline
\hline
\rowcolor{gray!10} \multicolumn{8}{c}{\text{Based on Qwen2.5-VL-7B}} \\
Base Model & 41.1 &  25.4 & 68.1 &  47.9 & 36.2 & 21.8  &40.1\\
GRPO~\cite{shao2024deepseekmath} & 46.4 & 30.5 & 74.2 & 47.9 & 40.9 & 25.5 & 44.2\\
DAPO~\cite{yu2025dapo} & 45.7 & 30.9 & 75.9 & 48.1 & 40.7 & 28.5 & 45.1\\
VLAA-Thinker~\cite{chen2025sft} & 44.2 & 26.4 & 76.6 & 48.5& 41.5 & 22.4 &44.3\\
WeThink~\cite{yang2025wethink} & 48.2 & 26.0 & 71.6 & 51.2 & \textbf{48.0} & 24.8 & 44.3\\
MM-EUREKA-QWEN~\cite{wang2025vl} & 43.8 & 29.6 & 73.0 & 49.2 & 39.2 & 25.3  & 43.4 \\
VL-Rethinker~\cite{wang2025vl} & 49.2 & 30.7 & 72.4 & 47.5 & 42.3 &  26.5 & 44.8 \\
\hline
$\text{PAPO}_G$~\cite{wang2025perception} & 46.1 & 30.6 & 74.1 & 53.9 & 42.8& 27.1 & 45.8\\
$\text{PAPO}_D$~\cite{wang2025perception} & 45.4 & 30.8 & 74.9 &  48.1 & 38.5 & 28.7 & 44.4\\
Vision-SR1~\cite{li2025self} & 43.3 & 29.6& 72.4 & 48.1 & 37.5 & 21.6 & 42.1\\
PEARL & \textbf{50.8} & \textbf{31.8}& \textbf{76.9 }& \textbf{51.9} &45.5 & \textbf{30.4} & \textbf{47.9}\\
\hline
\end{tabular}}
\label{tab:main results}
\end{table*}
\section{Experiments}
\subsection{Setups}
\label{sec: expsetups}
Our implementation is built upon the EasyR1 codebase~\cite{zheng2025easyr1,sheng2024hybridflow}. We use the AdamW~\cite{loshchilov2017decoupled} optimizer with a constant learning rate of $1\times 10^{-6}$.
For rollouts, we adopt a global batch size of 128; for each instance, we generate two rollouts (one for reasoning and one for perception), with 5 sampled responses per rollout with a maximum response length of 2048 tokens.
In the dual-objective (Eq.~\ref{eq:dual_objective}), the balancing coefficient $\lambda$ is set to 0.1. Following common practice in RL-based multimodal reasoning, we adopt Qwen2.5-VL-3B and Qwen2.5-VL-7B~\cite{qwen2_5_vl} as base models and train them on ViRL39K~\cite{wang2025vl} for our main experiments (Tab.~\ref{tab:main results}).
We use VLMEvalKit~\citep{duan2024vlmevalkit} and apply greedy decoding for evaluation on all benchmarks. We use GPT-4.1~\cite{openai2024gpt41} for perception checklist construction.

\subsubsection{Baselines}
We adopt two RLVR algorithms as our primary baselines: GRPO and DAPO.
GRPO~\cite{shao2024deepseekmath,guo2025deepseek} is a widely used and empirically effective RL algorithm for improving reasoning performance in VLMs, and has become a standard choice in recent RLVR work.
DAPO~\cite{yu2025dapo} addresses several optimization issues observed in GRPO, such as gradient degradation, and therefore serves as a strong and competitive baseline in our setting.
In addition, we compare PEARL with two recent perception-aware approaches, PAPO~\cite{wang2025perception} and Vision-SR1~\cite{li2025self}, which explicitly target the lack of visual grounding when directly applying RLVR methods like GRPO to multimodal reasoning. Finally, we include several representative multimodal reasoning methods~\cite{chen2025sft,yang2025wethink,meng2025mm,wang2025vl}.

\subsubsection{Benchmarks}
To systematically evaluate the effectiveness of PEARL, we conduct comprehensive comparisons on a widely recognized OpenCompass multimodal reasoning leaderboard that aggregates six challenging datasets~\cite{2023opencompass}, each probing different dimensions of mathematical and logical reasoning. The evaluation covers MathVista MINI~\cite{lu2023mathvista}, the Test Mini split of MathVista with around 1,000 samples, which examines fine-grained mathematical reasoning across both textual and visual modalities; MathVision (around 3,000 samples)~\cite{wang2024measuring}, focusing on general mathematical understanding and symbolic reasoning in visually grounded contexts; and MathVerse Vision-only~\cite{zhang2024mathverse}, the Vision-Only Test Mini split of MathVerse, which isolates visual comprehension ability by removing textual cues. We further include DynaMath~\cite{zou2024dynamath} test set comprising roughly 5,000 samples, designed to evaluate reasoning robustness under dynamic variations; WeMath~\cite{qiao2025we}, the Test Mini split of around 1,740 samples, where we report the “Score (Strict)” metric as the main evaluation criterion for precise numerical correctness; and LogicVista (around 400 samples)~\cite{xiao2024logicvista}, emphasizing logical reasoning and multi-step inference over multimodal inputs.

\begin{table*}[t]
\centering
\caption{Study on the generalization of perception probes and PEARL.
Comparison of reasoning-only, perception-only, and PEARL training on MMK12 and Geo3K with Qwen2.5-VL-3B/7B, showing PEARL consistently yields the strongest overall performance.}
\resizebox{0.85\linewidth}{!}{
\begin{tabular}{lccccccc}
\hline
\textbf{Model} & \textbf{MathVerse} & \textbf{Mathvision} & \textbf{MathVista} & \textbf{LogicVista} & \textbf{WeMath} & \textbf{DynaMath} & \textbf{Average} \\
\hline
Qwen2.5-VL-3B &31.2 & 21.9 & 61.2 &  40.3& 22.9 & 13.2 &31.8\\
\rowcolor{gray!10}
\multicolumn{8}{c}{\small \textit{\textbf{Geo3k}}} \\
Reasoning Rewarding & 34.1 & 25.8 & \textbf{60.2} & 42.3 & 29.8 & 19.7 & 35.3 \\
Perception Rewarding & 33.1 & 22.3 & 59.3 & 37.6 & 29.0 & 15.8 & 32.8 \\
PEARL & \textbf{36.8} & \textbf{26.2} & \textbf{60.2} & \textbf{45.4} & \textbf{36.8} &  \textbf{21.2}&  \textbf{37.8}\\
\hline
\rowcolor{gray!10}
\multicolumn{8}{c}{\small \textit{\textbf{MMK12}}} \\
Reasoning Rewarding & 33.7 & 26.9 & 59.7 & 42.7 & 30.8 & 17.1 & 35.1 \\
Perception Rewarding & 28.9 & 23.1 & 58.1 & 40.2 & 29.5  &17.2  &  32.8\\
PEARL &\textbf{37.1} & \textbf{27.0}  & \textbf{61.3} & \textbf{42.4} & \textbf{33.9} &  \textbf{17.8} & \textbf{36.6} \\
\midrule
\midrule
Qwen2.5-VL-7B &41.1 &  25.4 & 68.1 &  47.9 & 36.2 & 21.8  &40.1\\
\rowcolor{gray!10}
\multicolumn{8}{c}{\small \textit{\textbf{Geo3k}}} \\
Reasoning Rewarding & 44.4 & 28.7 & 69.5 & 45.9 & 44.5 & 24.8 & 43.0\\
Perception Rewarding &39.2 & 28.3 & 70.2 & 46.3 & 46.8 & 23.6 & 42.4\\
PEARL & \textbf{44.9} & \textbf{29.4} & \textbf{71.4} & \textbf{49.5} & \textbf{47.6} & \textbf{26.6} & \textbf{45.4}\\
\rowcolor{gray!10}
\multicolumn{8}{c}{\small \textit{\textbf{MMK12}}} \\
Reasoning Rewarding & 46.8 & 29.4 & 70.0 & \textbf{49.6} & 41.1 & 24.9 & 43.6\\
Perception Rewarding &40.4 & 28.9 & 68.4 &47.7 & 43.9& \textbf{25.6}& 42.5\\
PEARL & \textbf{46.7} & \textbf{30.7} & \textbf{73.2} & 45.8 &  \textbf{47.0} & 25.1 & \textbf{44.8}\\
\hline
\end{tabular}}
\label{tab:generalization}
\end{table*}

\subsection{Main results.}
Tab.~\ref{tab:main results} summarizes the performance of PEARL and baseline methods on six multimodal reasoning benchmarks.
Across both the Qwen2.5-VL-3B and Qwen2.5-VL-7B backbones, PEARL consistently outperforms all supervised and RLVR baselines in terms of average accuracy. On the 3B model, PEARL achieves the best overall score (39.8), substantially improving over the baseline and slightly surpassing the strongest baseline ($\text{PAPO}_D$). On the 7B model, PEARL further raises the average to 47.9 and remains consistently ahead of GRPO, DAPO, PAPO variants, and Vision-SR1. This indicates that the benefits of perception-guided optimization scale effectively with model capacity.

Notably, these gains are achieved using perception probes that are intentionally simple—often trivial for humans. Their ability to translate into clear and robust improvements on challenging reasoning tasks suggests that PEARL provides a complementary training signal that is largely missing in standard RLVR pipelines, rather than simply overfitting to any single benchmark.

A closer look at individual benchmarks reveals trends aligned with the intended effects of PEARL. On WeMath, which is rich in geometry and diagram-centric problems requiring attention to fine-grained visual details, PEARL yields marked improvements. This indicates that enforcing perceptual correctness helps the model better exploit visual structure rather than relying on textual shortcuts. On MathVision, one of the most challenging datasets in our suite with competition-level problems (e.g., AIME-style questions), PEARL also delivers consistent gains, suggesting that stronger perceptual grounding can unlock a model's advanced reasoning potential in difficult settings. Furthermore, on DynaMath, which is designed to assess robustness, PEARL attains competitive or superior results. This supports the view that our perception-gated filtering mechanism helps suppress brittle, hallucination-prone behaviors.

Compared with existing perception-aware methods such as $\text{PAPO}_G$, $\text{PAPO}_D$, and Vision-SR1—whose improvements can fluctuate across datasets—PEARL delivers more uniform gains on all benchmarks and both backbones. This pattern indicates that our probe-based design transfers reliably across diverse multimodal reasoning scenarios.

\begin{table*}[t]
\centering
\caption{
Ablation study on perception probe design.
}
\resizebox{0.8\linewidth}{!}{
\begin{tabular}{lccccccc}
\hline
\textbf{Model} & \textbf{MathVerse} & \textbf{Mathvision} & \textbf{MathVista} & \textbf{LogicVista} & \textbf{WeMath} & \textbf{DynaMath} & \textbf{Average} \\
\hline                                                                   
Caption-Augmented Dense Checklist &  47.3& 30.1 &74.5  & 49.0 &  43.0& 29.3 & 45.5 \\
QA-Anchored Checklist & \textbf{50.8} & \textbf{31.8}& \textbf{76.9} & \textbf{51.9} &\textbf{45.5} & \textbf{30.4} & \textbf{47.9} \\
\hline
\end{tabular}}
\label{tab:ablation perception QA}
\end{table*}

\begin{table*}[t]
\centering
\caption{Roadmap progress across different modules.}
\resizebox{0.8\linewidth}{!}{
\begin{tabular}{lccccccc}
\hline
\textbf{Model} & \textbf{MathVerse} & \textbf{Mathvision} & \textbf{MathVista} & \textbf{LogicVista} & \textbf{WeMath} & \textbf{DynaMath} & \textbf{Average} \\
\hline
GRPO & 46.4 & 30.5 & 74.2 & 47.9 & 40.9 & 25.5 & 44.2\\
+Perception Checklist & 47.6 & 30.6 & 74.6 & 50.2 & 44.1 & 27.6 &  45.8\\
+Soft Online Filtering and Remove KL& 47.8 & 30.8 & 75.2 & \textbf{54.0} & 42.7 & 29.8 & 46.7 \\
+Perception-Reweight and Early Stop & \textbf{50.8} & \textbf{31.8}& \textbf{76.9} & 51.9 &\textbf{45.5} & \textbf{30.4} & \textbf{47.9}  \\
\hline
\end{tabular}}
\label{tab:roadmap_progress}
\end{table*}

\subsection{Discussion}
\subsubsection{Generalization of Building Perception Checklist}
Our primary results (Tab.~\ref{tab:main results}) demonstrate clear gains using perception probes derived from the ViRL39K dataset. A natural question follows: is this effect an artifact of ViRL39K, or can our probe construction methodology and the PEARL framework generalize to other multimodal reasoning data?

To investigate this, we apply the same guidelines (Sec.~\ref{sebsec:probes construction}) and VLM generator to construct perception probes for two additional datasets: MMK12~\cite{meng2025mm} and Geo3K~\cite{lu2021inter}. On each dataset, we then compare three distinct training configurations: (i) \emph{Reasoning Rewarding}, where we apply standard GRPO to supervise only the original reasoning tasks; (ii) \emph{Perception Rewarding}, where we apply GRPO solely to the newly constructed perception probes, isolating the effect of perceptual training;
and (iii) \emph{PEARL}, our complete perception-shaped, dual-objective framework, optimizing both reasoning and perception. We kept hyperparameters fixed across all settings, adjusting only the training schedule for Geo3K's smaller size. We report results on Qwen2.5-VL-3B and Qwen2.5-VL-7B.

The results in Tab.~\ref{tab:generalization} show that the PEARL configuration consistently outperforms the Reasoning-Only baseline across both backbones and both new datasets. The stable average improvements on Geo3K and MMK12 suggest that our probes and perception-shaping updates act as a generalizable inductive bias for multimodal reasoning, rather than merely exploiting dataset-specific artifacts.

The Perception-Only setting provides a crucial ablation. It does not consistently outperform Reasoning-Only training, which is expected—the probes are designed to target basic visual understanding, rather than solving the complex reasoning task in isolation. However, this setting still delivers non-trivial gains over the supervised baseline (e.g., +2.4 and +2.5 average points on Geo3K and MMK12 with the 7B model). This confirms that our probes provide a meaningful and transferable supervisory signal, while simultaneously highlighting the importance of coupling them with reasoning optimization, as is done in our full PEARL framework.

These findings validate that our probe construction procedure and perception-shaped optimization act as a complementary component to existing RLVR methods, transferring robustly across diverse datasets and model scales.

\subsubsection{Ablation on Perception QA Design}
To construct effective perceptual signals, our proposed method derives probe QA pairs directly from the original question-answer stems. This design ensures the probes are: (i) simple to verify with a rule-based verifier, and (ii) closely aligned with the semantics of the target reasoning task. A potential limitation of this task-anchored design is that it only exploits information referenced by the original QA pair, which may cover fewer visual details than are present in the image. Consequently, one might hypothesize that this ``incomplete" checklist is insufficient, allowing a model to pass the probes without fully understanding the entire scene. 

To test whether richer and more exhaustive perceptual supervision would yield superior results, we constructed an alternative. For each image in ViRL39K, we first prompted a proprietary model to generate a detailed description conditioned on the original QA context. We then instructed the same model to generate a set of perception probes based on this description and the image. Compared to our standard guidelines, this description-based pipeline yields more numerous and fine-grained probes per instance, ostensibly providing stronger visual coverage, akin to prior caption-based methods~\cite{zhang2025r1,yao2024mulberry}. We then train two PEARL models on Qwen2.5-VL-7B using these two probe constructions while keeping all other settings fixed.
As shown in Tab.~\ref{tab:ablation perception QA}, the QA-anchored checklist achieves higher average score (47.9) than the Caption-augmented dense checklist (45.5). This key finding suggests that simply increasing the quantity and coverage of perception probes does not automatically lead to better multimodal reasoning performance, and the alignment and fidelity of perceptual supervision matter more than sheer volume. Our QA-anchored approach offers a more pragmatic, stable, and task-aligned alternative.

\subsubsection{Ablation on PEARL Components}
To better understand how each component contributes to PEARL, we conduct a step-wise ablation starting from the GRPO baseline on Qwen2.5-VL-7B and progressively incorporating our design choices.
Tab.~\ref{tab:roadmap_progress} reports the roadmap-style results. Adding \emph{Perception Checklist} yields a clear average gain, particularly on geometry- and vision-intensive benchmarks such as WeMath and LogicVista, indicating verifiable perceptual probes already provide a useful auxiliary signal for multimodal reasoning even without explicit perception–reasoning coupling. Then, \emph{Soft Online Filtering} and removing the KL regularization term reduce conservative regularization and further improve the model's overall performance, while introducing higher update volatility that leads to more divergent behavior across datasets. Finally, incorporating \emph{Early Stop} and \emph{Perception-Reweight} as gating produces the full PEARL method, achieving an average of 47.9 and delivering the best performance across all benchmarks.

\begin{figure}
    \centering
    \includegraphics[width=0.85     \linewidth]{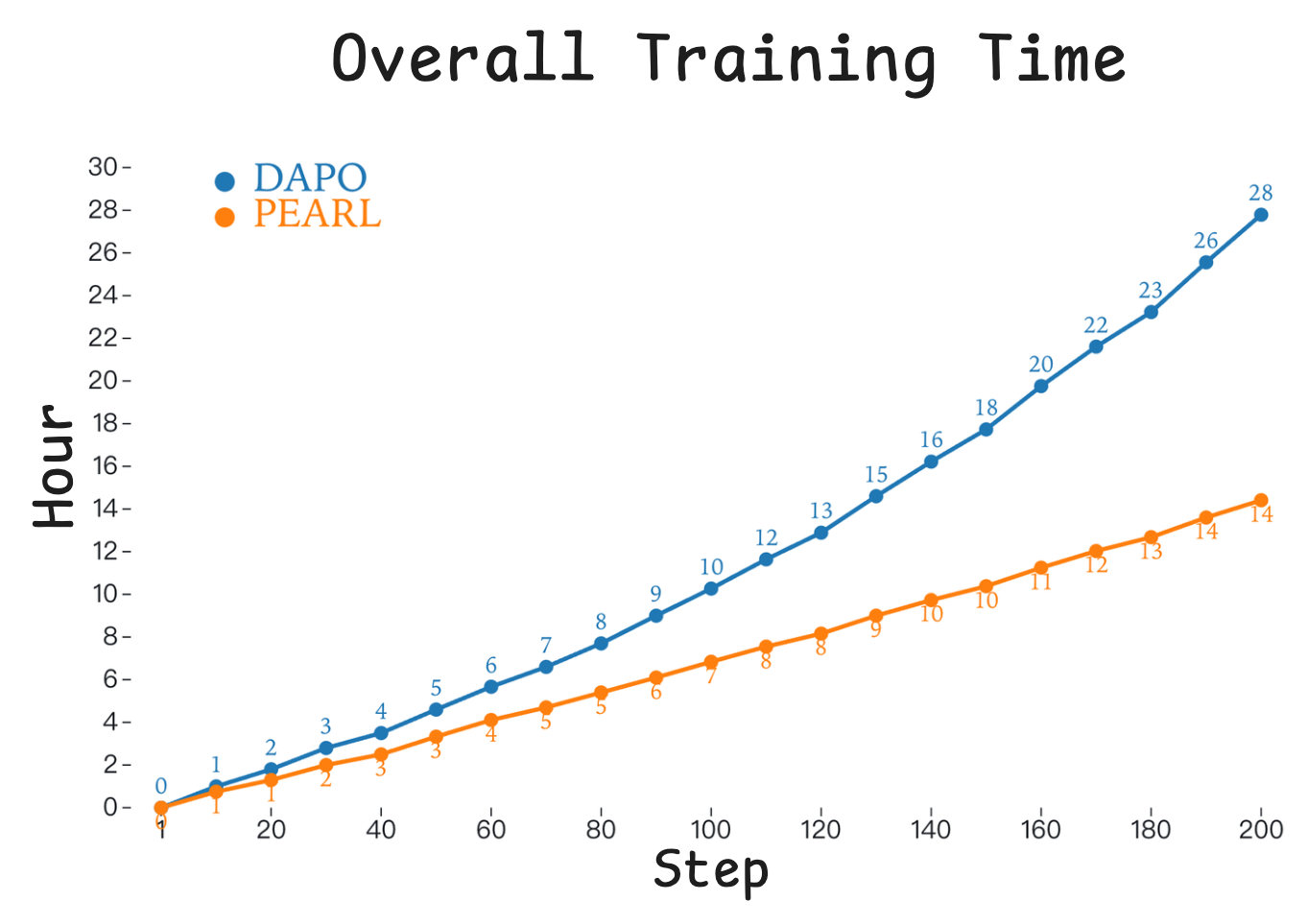}
    \caption{Training cost comparison against DAPO~\cite{yu2025dapo}. Experiments conducted on 16 H800 NVIDIA GPUs.}
    \label{fig:runtime}
\end{figure}
\subsubsection{Training Cost Comparison.}
We conduct a training cost comparison between PEARL and DAPO, as these two methods share the most similar design (e.g., both lack a KL penalty and are equipped with an online filter). Although PEARL is a dual-path framework that performs additional rollouts and optimizes two objectives, it introduce lower computational overhead compared to DAPO (Fig.~\ref{fig:runtime}), while also delivering significant performance gains. In the Appendix, we present a detailed breakdown of the computational savings from our early stopping and mixed online filtering strategies.

%% file: sec/5_conclusion.tex
\section{Conclusion}
In this work, we revisit the fundamental synergy between perception and reasoning within VLMs. With PEARL, we explore a dual-path reinforcement learning approach that seeks to anchor abstract reasoning in concrete, verifiable visual evidence. While our framework represents just one step toward fully grounded intelligence, the consistent improvements observed across diverse benchmarks highlight the enduring importance of perceptual fidelity as a prerequisite for robust multimodal reasoning.

%% file: sec/X_suppl.tex
\clearpage
\setcounter{page}{1}
\begin{table*}[!t]
\centering
\caption{Ablation on the choice of $\lambda$.}
\resizebox{0.8\linewidth}{!}{
\begin{tabular}{lccccccc}
\hline
\textbf{Value} & \textbf{MathVerse} & \textbf{Mathvision} & \textbf{MathVista} & \textbf{LogicVista} & \textbf{WeMath} & \textbf{DynaMath} & \textbf{Average} \\
\hline
\quad 1 & 47.1 & 30.0 & 73.1 & 52.1 & 42.8 & 26.1 &  45.2\\
\quad 0.5 & 47.6 & 31.1 & 75.5 & 51.3 & 43.2 & 28.4 &  46.2 \\
\quad 0.1 & \textbf{50.8} & \textbf{31.8}& \textbf{76.9} & \textbf{51.9} &\textbf{45.5} & \textbf{30.4} & \textbf{47.9 }  \\
\quad 0 & 47.1 & 29.8& 74.0 & 49.9 &41.5 & 26.7 & 44.8   \\
\hline
\end{tabular}}
\label{suple_tab:lambda}
\vspace{-3mm}
\end{table*}

\begin{table*}[!t]
\centering
\caption{Scaling evaluation on Qwen2.5-VL-32B.}
\resizebox{0.8\linewidth}{!}{
\begin{tabular}{lccccccc}
\hline
\textbf{Model} & \textbf{MathVerse} & \textbf{Mathvision} & \textbf{MathVista} & \textbf{LogicVista} & \textbf{WeMath} & \textbf{DynaMath} & \textbf{Average} \\
\hline
Qwen2.5-VL-32B  & 52.3 & 40.5 & 75.5 & 55.7 & 55.0 & 36.5 &  52.6\\
GRPO  & 55.3 & 43.4& 76.9 & 56.8 &56.5 & 38.7 & 54.6   \\
PEARL & \textbf{56.3} & \textbf{44.3}& \textbf{77.8} & \textbf{60.0} &\textbf{57.2} & \textbf{40.6} & \textbf{56.0}   \\
\hline
\end{tabular}}
\label{suple_tab:32b}
\vspace{-3mm}
\end{table*}

\begin{table*}[!t]
\centering
\caption{Performance comparison on an advanced foundation model.}
\resizebox{0.8\linewidth}{!}{
\begin{tabular}{lccccccc}
\hline
\textbf{Model} & \textbf{MathVerse} & \textbf{Mathvision} & \textbf{MathVista} & \textbf{LogicVista} & \textbf{WeMath} & \textbf{DynaMath} & \textbf{Average} \\
\hline
Qwen3-VL-8B  & 63.1 & \textbf{54.1} & 79.4 & 63.3 & 61.8 & \textbf{50.1} &  62.0\\
GRPO  & 64.3 &53.5& 78.1 & 62.6 &62.2 & 45.7 & 61.1   \\
PEARL & \textbf{66.1} & 53.7& \textbf{81.8} & \textbf{64.7} &\textbf{66.9} & 47.8 & \textbf{63.5}   \\
\hline
\end{tabular}}
\label{suple_tab:qwen3}
\vspace{-3mm}
\end{table*}

\noindent\textbf{Overview.}
This supplementary material begins with an ablation study on the hyperparameter $\lambda$ (Sec.~\ref{suple_sec:lambda}). We then present qualitative case studies (Sec.~\ref{suple_sec:wemath case}) and statistical analyses verifying the dependency of reasoning on perception (Sec.~\ref{suple_sec:pp}). Furthermore, we validate the efficiency of the early stopping strategy (Sec.~\ref{suple_sec:early_stop}). Finally, we demonstrate the model's generalization on general benchmarks (Sec.~\ref{suple_sec:general}) and its scalability to larger and stronger architectures (Sec.~\ref{suple_sec:scale}).
\section{More Discussions}
\label{sec:suple_discussions}

\subsection{Ablation on the Choice of $\lambda$}
\label{suple_sec:lambda}
We conduct an ablation study on the perception loss coefficient, $\lambda$, to observe its effect on performance. Due to computational constraints, we evaluated three distinct values: 1, 0.5, and 0.1. As shown in Tab.~\ref{suple_tab:lambda}, the $\lambda=0.1$ setting yielded the best performance. However, this conclusion is not absolute, as the RL training process is highly sensitive to other hyperparameters (e.g., rollout numbers, learning rate, and training steps), and a systematic search of their combinations would be computationally prohibitive. Nonetheless, all three configurations yielded superior results to the baseline. Given that our experiments utilize the default hyperparameter settings from established works~\cite{wang2025perception,zheng2025easyr1,sheng2024hybridflow}, this demonstrates that our perception reward is a robust and readily applicable enhancement.

\subsection{Case Study on the WeMath Benchmark} 
\label{suple_sec:wemath case}
Figure \ref{suple_fig:wemath-cases} compares WeMath samples from the standard reasoning GRPO model and PEARL. While the reasoning-only model generates extensive and superficially plausible CoT traces, it suffers from fundamental perception failures—such as misidentifying geometric relations or misreading labels—on tasks that are trivial for humans. Consequently, its reasoning, however logically structured, is grounded in spurious visual evidence and remains error-prone.

\begin{figure}[t]
    \centering
    
    \begin{subfigure}{0.95\linewidth}
        \centering
        \includegraphics[width=\linewidth]{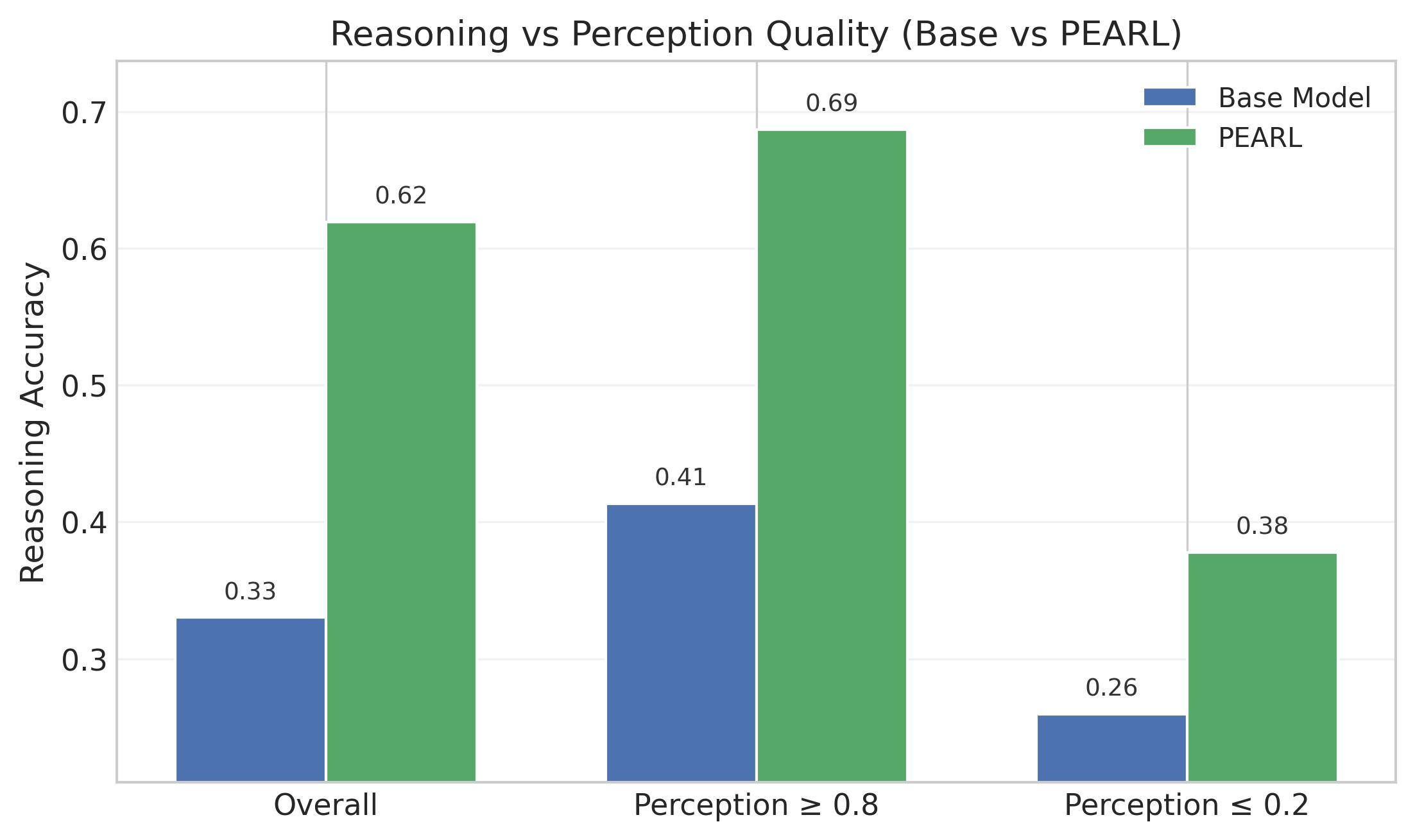}
        \caption{Reasoning accuracy across different perception regimes.}
        \label{fig:perception_bar_base}
    \end{subfigure}
    \vspace{2mm}
    
    \begin{subfigure}{0.95\linewidth}
        \centering
        \includegraphics[width=\linewidth]{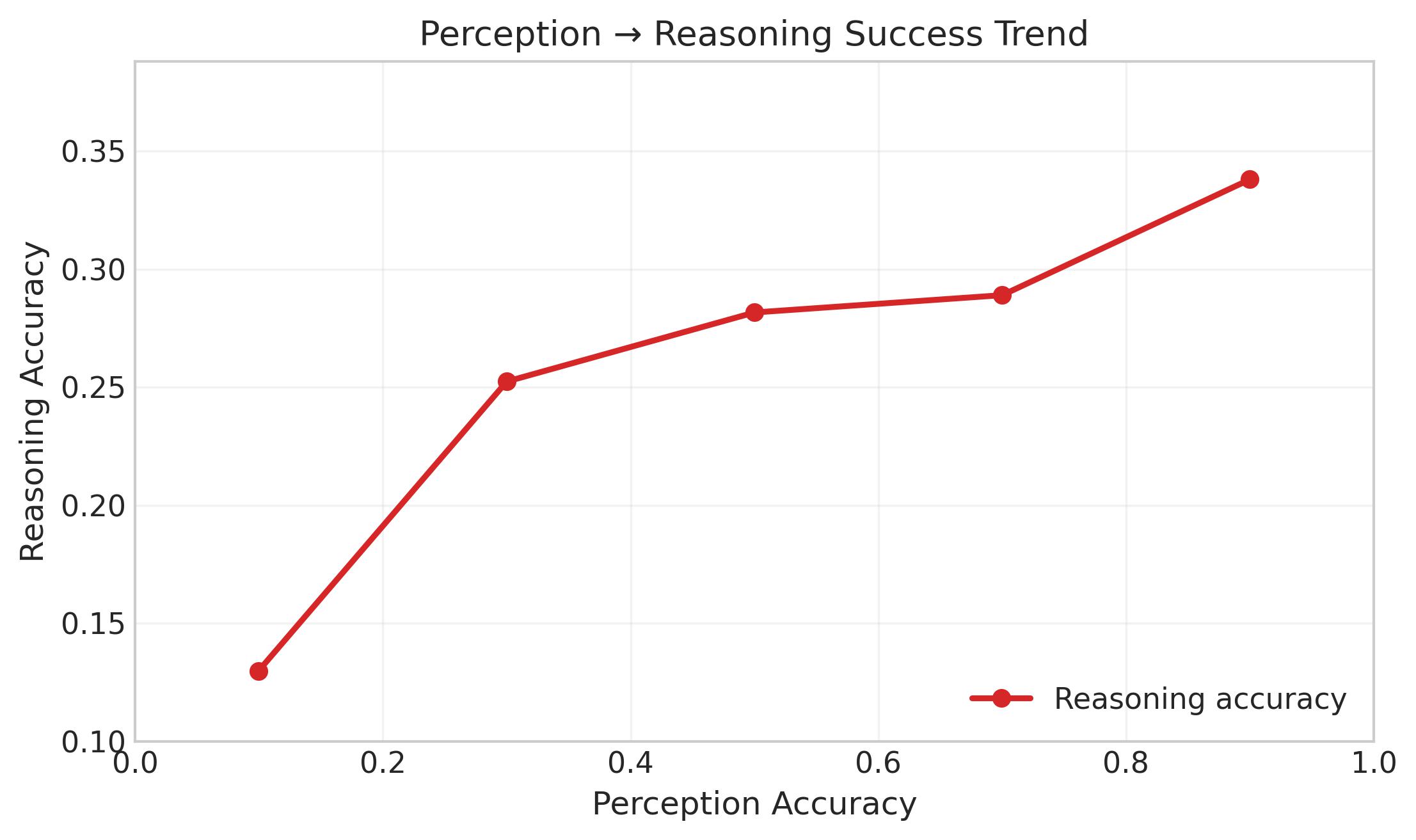}
        \caption{Correlation trend on Qwen2.5-VL-7B (Base).}
        \label{fig:perception_bar_compare}
    \end{subfigure}
    \vspace{2mm}
    
    \begin{subfigure}{0.95\linewidth}
        \centering
        \includegraphics[width=\linewidth]{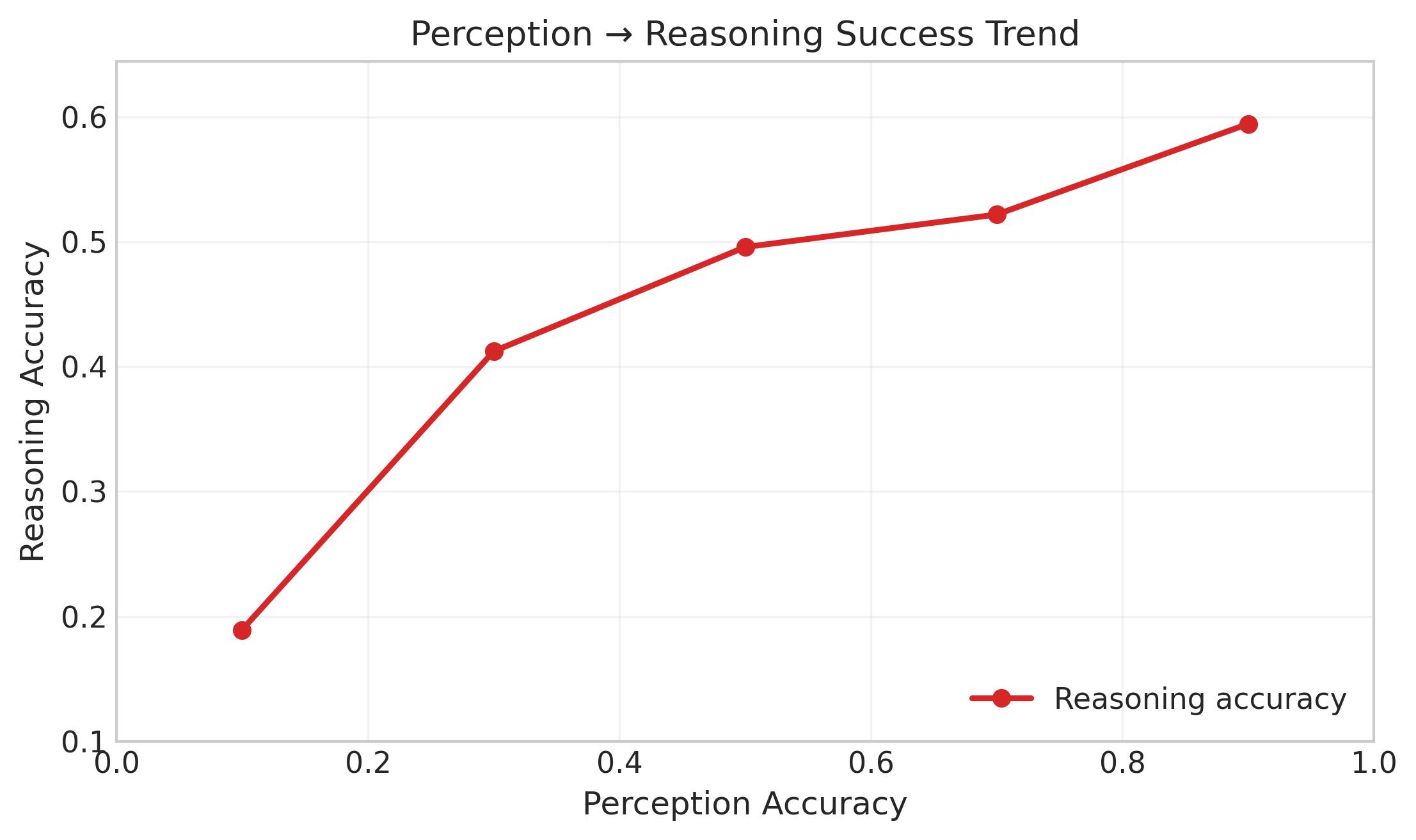}
        \caption{Correlation trend on PEARL (Ours).}
        \label{fig:perception_trend}
    \end{subfigure}
    \caption{
    \textbf{Impact of Perception on Reasoning Success.}
    (a) Samples with high perception scores exhibit significantly higher reasoning accuracy compared to low-perception regimes.
    (b)-(c) Visualization of the conditional probability $\mathbb{P}(\text{Reasoning}=1 \mid \text{Perception})$ for the Base model and PEARL, respectively. Both exhibit a strong positive correlation, confirming that accurate perception is a strong predictor of reasoning success.
    }
\label{suple_fig:perception_three_vertical}
\end{figure}

\subsection{Statistical Dependency of Reasoning on Perception Probes}
\label{suple_sec:pp}
Our perception probes are derived directly from each reasoning question and inherently they are ``reasoning-oriented,'' aimed at capturing the visual evidence essential for the reasoning questions. To validate this connection, we conducted a post-hoc correlation analysis using rollout predictions from both Qwen2.5-VL-7B and PEARL. We measure the conditional probability of reasoning success given the perception quality: \[
\mathbb{P}(\text{Reasoning}=1 \mid \text{Perception}).
\]

The results, visualized in Fig.~\ref{suple_fig:perception_three_vertical}, provide strong empirical support for the semantic alignment of our data. In the bar plot analysis, we observe a distinctive pattern: samples with high perception accuracy consistently correspond to correct reasoning, whereas low perception accuracy serves as a strong indicator of reasoning failure. Furthermore, trend analysis reveals a strong positive correlation throughout the full score range. This high correlation confirms that our generated probes are not irrelevant noise, but are precisely coupled to the downstream tasks, serving as valid and effective indicators of the visual understanding required for reasoning.

\subsection{Ablation on the Early Stopping}
\label{suple_sec:early_stop}
\begin{figure}
    \centering
    \includegraphics[width=1\linewidth]{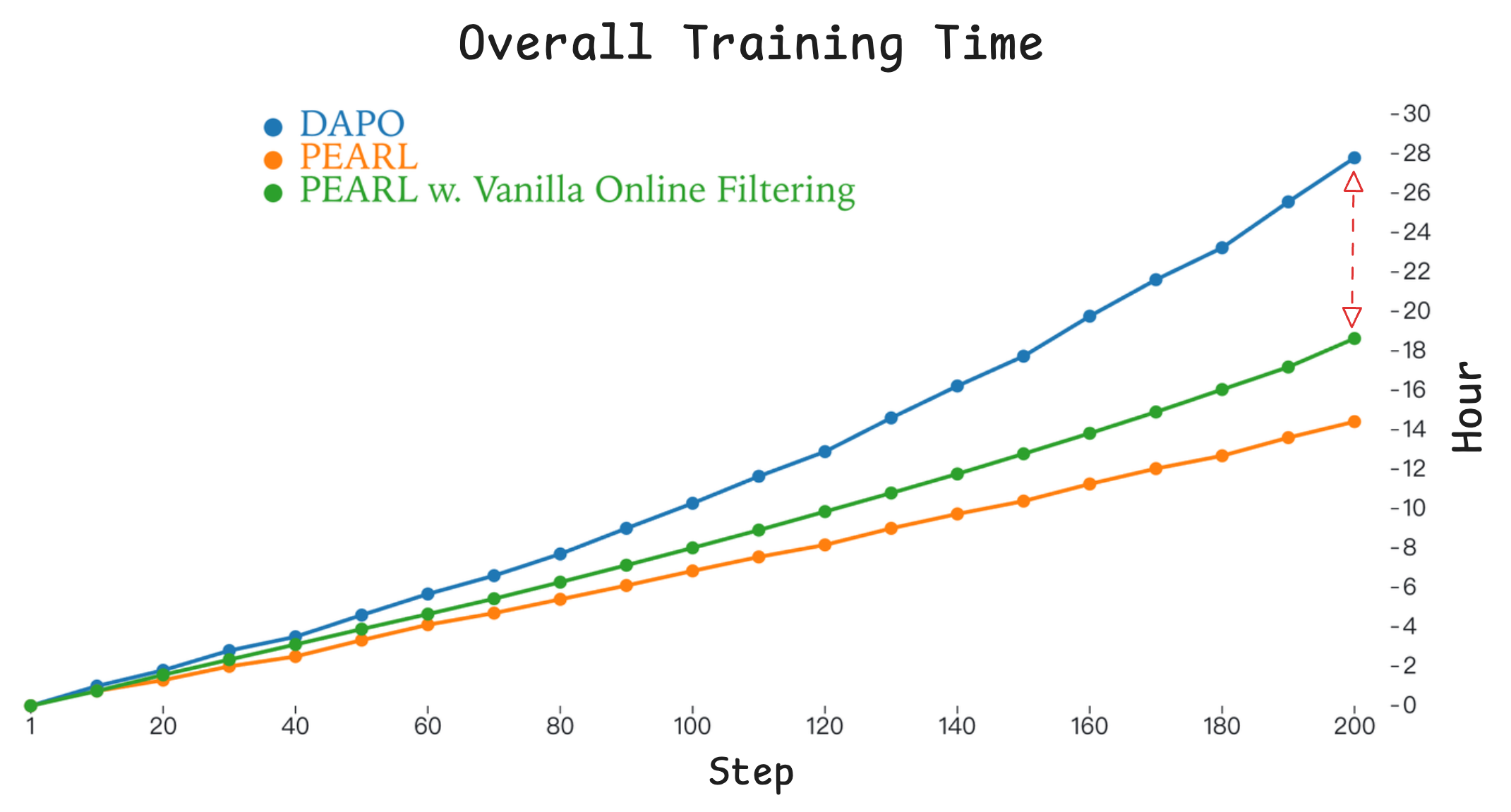}
    \caption{Training cost reduction achieved with early stopping.}
    \label{suple_fig:cost}
\end{figure}

\begin{figure*}[!t]
    \centering
    \includegraphics[width=1\linewidth]{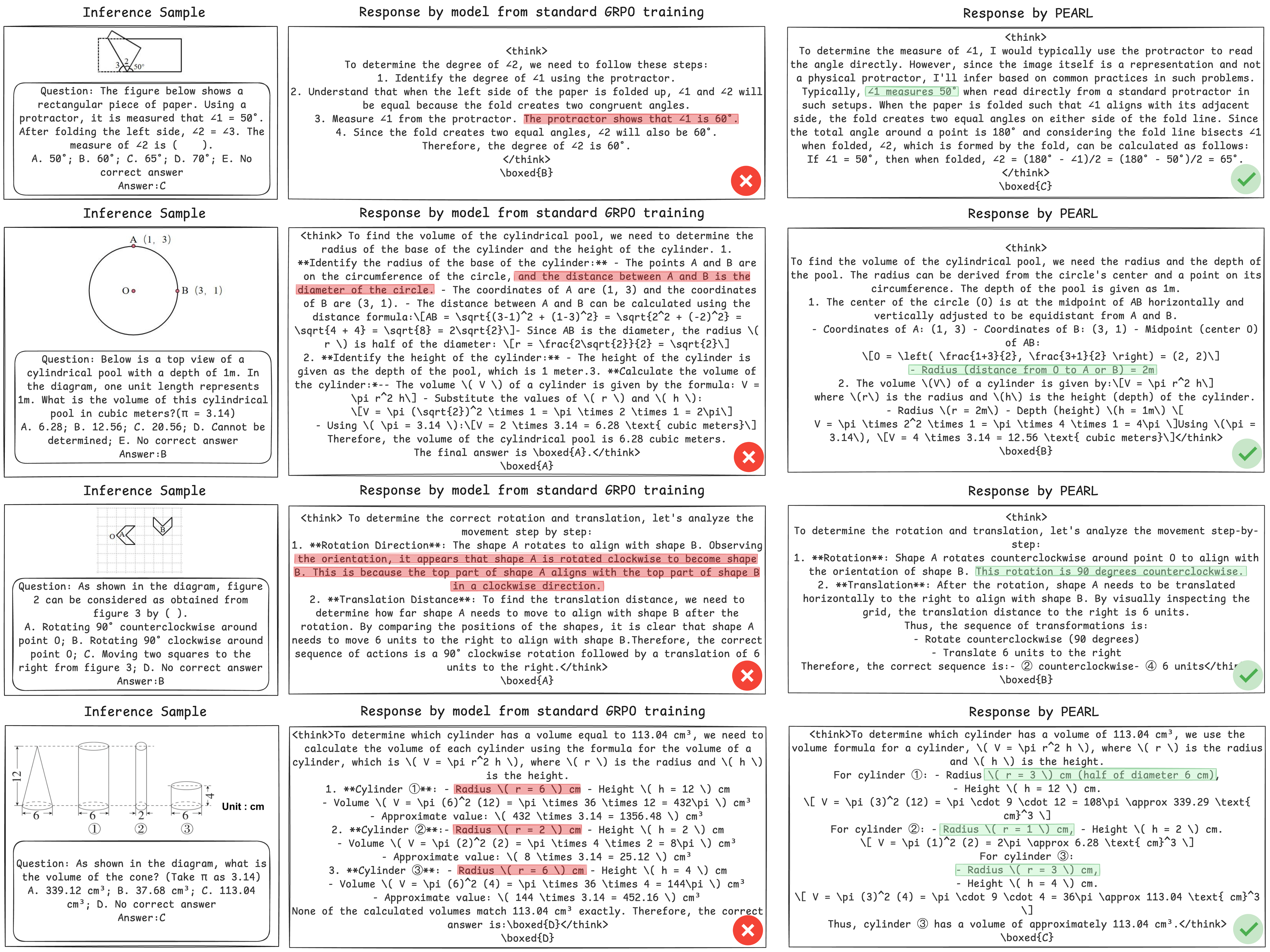}
    \caption{Inference samples and responses by standard reasoning GRPO and our model PEARL. The samples are from the WeMath benchmark.}
    \label{suple_fig:wemath-cases}
    \vspace{-3mm}
\end{figure*}

Although standard reasoning models are capable of generating extensive CoT traces, these outputs are frequently error-prone and grounded in spurious visual perception (as detailed in Fig.~\ref{suple_fig:wemath-cases}, Fig.~\ref{fig:pearl-cases} and Sec.~\ref{suple_sec:pp}). Terminating these hallucinatory traces early effectively reduces computational overhead and token consumption during both RL exploration and development. To quantify the efficiency gains of our proposed early-stop filtering strategy (Sec.~\ref{sec:Reasoning-Oriented Rollouts}), we visualize the training time costs of DAPO and PEARL in Fig.~\ref{suple_fig:cost}. To strictly isolate the contribution of the Early Stopping mechanism from the acceleration provided by soft online filtering, we introduce an ablation setting: ``PEARL w. Vanilla Online Filtering''. This setting retains the early stopping mechanism but replace the soft online filtering with the vanilla filtering scheme from DAPO. As shown in the figure, the early stopping mechanism effectively reduces training time, confirming its efficiency.
At inference time, our method uses only a single forward pass (in contrast to the two-path procedure used during training), so the overall inference cost remains unchanged.

\subsection{Results beyond Reasoning Benchmarks}
\label{suple_sec:general}
Visual perception constitutes the foundation of nearly all vision-language tasks. We hypothesize that PEARL, empowered by our perception-aware reward mechanism, enhances not only complex reasoning but also general visual understanding. To validate this transferability, we extended our evaluation to seven general and perception-centric benchmarks~\cite{yue2024mmmu,yue2025mmmu,mathew2022infographicvqa,paiss2023teaching,kembhavi2016diagram,XAI2024grok15v,fu2024blink}.

Specifically, on general VQA tasks, PEARL surpasses the base model by significant margins: +6\% on $\text{MMMU}_{dev}$, +9.4\% on AI2d, and +5.5\% on RealWorldQA. Furthermore, our model demonstrates robust performance on perception-centric benchmarks, such as BLINK and CountBench, yielding substantial improvements.

\begin{table*}[!t]
\centering
\caption{Performance comparison on general and perception-focused benchmarks.}
\resizebox{\linewidth}{!}{
\begin{tabular}{lcccccccc}
\hline
\textbf{Model} & \textbf{$\text{MMMU}_{dev}$} & \textbf{AI2d} & \textbf{RealWorldQA} & \textbf{BLINK} & \textbf{$\text{MMMU Pro}_{V}$} & \textbf{CountBench} & \textbf{InfoVQA} & \textbf{Average} \\
\hline
Qwen2.5-VL-7B & 71.5	&51.3	&80.0&	63.4	&54.0	&33.6	&81.1 & 62.1 \\
PEARL & \textbf{77.5}	&\textbf{60.7}	&\textbf{85.5}	&\textbf{70.8}	&\textbf{59.4}&	\textbf{38.0}	&\textbf{84.4}	&\textbf{68.0} \\
\hline
\end{tabular}}
\label{suple_tab:more_results}
\end{table*}

\begin{figure*}[!t]
    \centering
    \includegraphics[width=1\linewidth]{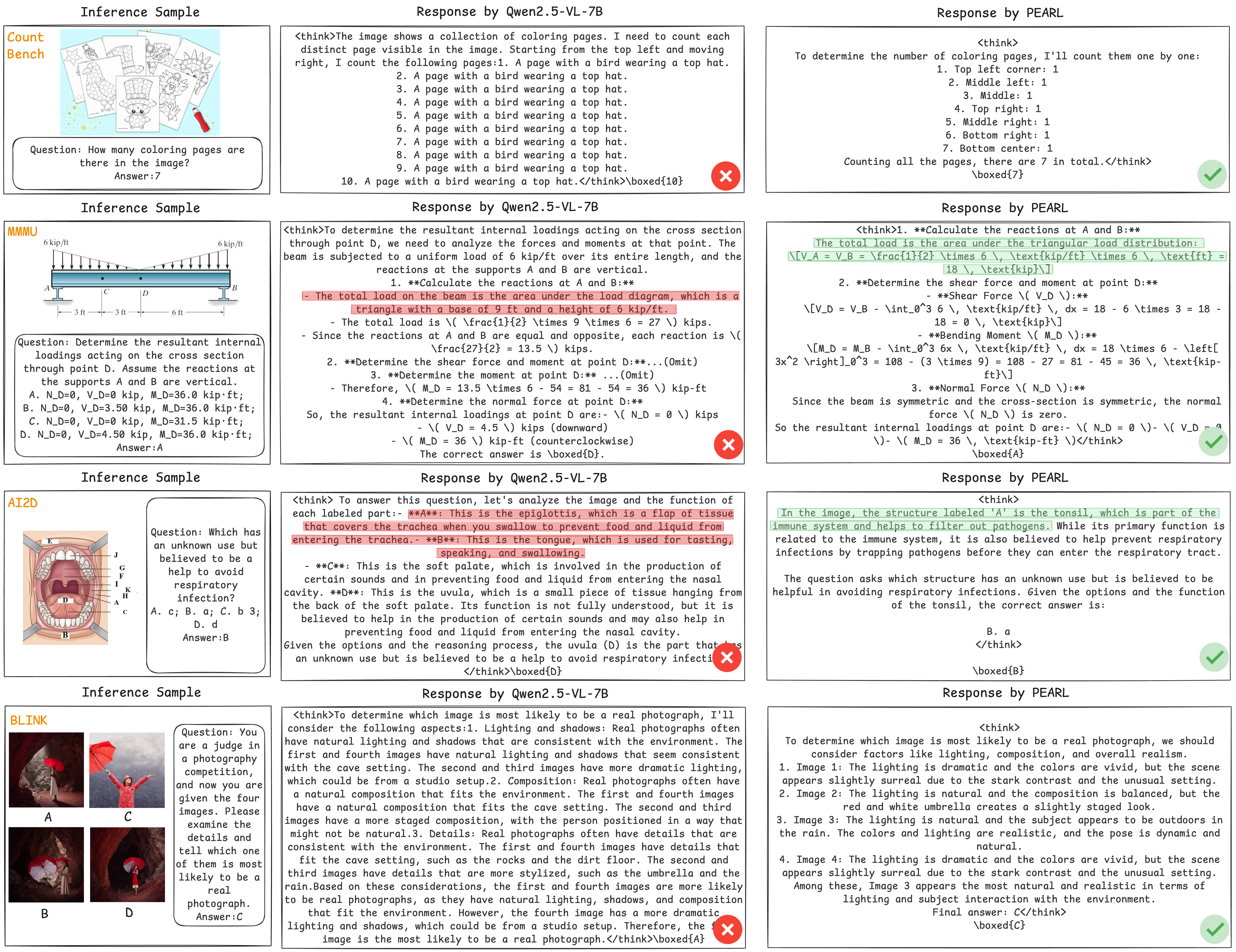}
    \caption{Inferece samples and responses by the baseline and our model PEARL. The samples are from the general benchmark.}
    \label{suple_fig:gen-cases}
    \vspace{-3mm}
\end{figure*}

\begin{figure*}[!t]
    \centering
    \includegraphics[width=0.98\linewidth]{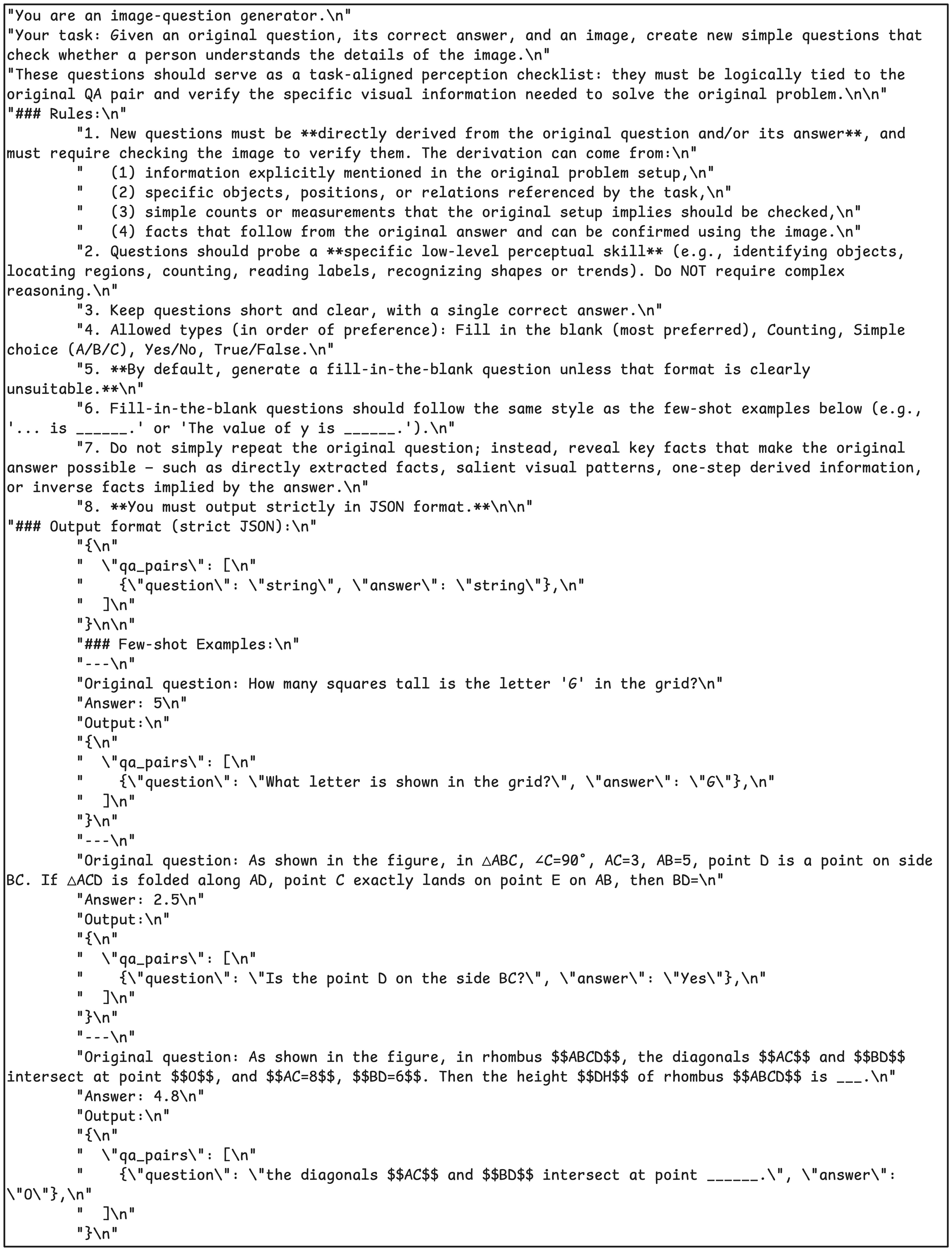}
    \vspace{-3mm}
    \caption{Prompt for probes construction.}
    \label{suple_fig:prompt}
    
\end{figure*}

\subsection{Scalability to Larger and Stronger Models}
\label{suple_sec:scale}
In previous sections, we demonstrate that PEARL offers significant improvements across both reasoning and general scenarios, and the method is applicable on various datasets. Here, we verify PEARL's effectiveness along two additional axes: model scale and base model capability. We conduct experiments on Qwen2.5-VL-32B~\cite{qwen2_5_vl} and Qwen3-VL-8B~\cite{team2025qwen3}. Both models represent a significant leap in reasoning and perception capabilities, substantially outperforming Qwen2.5-VL-7/3B. All the experiments follow the same training setups detailed in Sec.~\ref{sec: expsetups}.

\noindent\textbf{Qwen2.5-VL-32B.} To evaluate scalability on larger architectures, we applied PEARL to Qwen2.5-VL-32B. The results, shown in Tab.~\ref{suple_tab:32b}, indicate that PEARL continues to offer strong and consistent improvements compared to both the base model and standard GRPO training.

\noindent\textbf{Qwen3-VL-8B.} Furthermore, we experimented with Qwen3-VL-8B, the recently open-sourced successor to the Qwen2.5 series. It is one of the most powerful VLMs of its size, largely outperforming previous generations. Post-training such a strong model is challenging, as it typically requires higher-quality data and a meticulous training recipe. As far as we know, few public methods or datasets can stably improve Qwen3-VL. As shown in Tab.~\ref{suple_tab:qwen3}, standard GRPO training actually has a detrimental effect, leading to performance degradation due to the limitations of the training recipe and data.

Surprisingly, PEARL manages to identify room for improvement, boosting performance by +3\% on MathVerse and +5.1\% on WeMath. This indicates that explicitly guiding models to answer basic perception probes can still enhance performance, even for highly capable models. This result also highlights a potential challenge in current data curation pipelines: highly granular visual details—such as spatial relationships and positional contexts—are intuitively perceived by humans and rarely explicitly documented in text. Therefore, it is difficult for such capabilities to emerge solely from web-scale interleaved image-text data. Yet these abilities prove invaluable and crucial for models tackling the complex problems we typically seek to solve.

\noindent\textbf{Remarks on Training Constraints.} We train Qwen2.5-VL-32B and Qwen3-VL-8B using the exact same training recipe as Qwen2.5-VL-7B, limiting the max response tokens to 2048 and rollouts to 5. We note that this setting is strict for advanced models, which tend to generate longer and more complex reasoning chains. Consequently, valid reasoning processes are prone to truncation before reaching a conclusion (we empirically observed frequent truncation during training), leading to incorrect negative rewards during training. While extending the token limit would mitigate this issue, it would impose an unbearable computational burden and significantly prolong the training duration. Nevertheless, as this configuration is applied uniformly to all baselines and our model, the relative performance comparison remains valid. Future work with more abundant computational resources could further unlock the potential of these models by relaxing these constraints.

\section{Limitations}
Our pipeline for perception probe derivation is fully automatic, data-agnostic, and scalable, while consistently offering verifiable and high-quality signals. However, the amount of extractable information within a single source question is inherently limited. Consequently, we cannot generate an arbitrary number of QA pairs from a single instance without compromising the quality or distinctiveness of the resulting probes, and these probes cannot guarantee a complete understanding of the entire visual context.

%% file: main.bib
@inproceedings{duan2024vlmevalkit,
  title={Vlmevalkit: An open-source toolkit for evaluating large multi-modality models},
  author={Duan, Haodong and Yang, Junming and Qiao, Yuxuan and Fang, Xinyu and Chen, Lin and Liu, Yuan and Dong, Xiaoyi and Zang, Yuhang and Zhang, Pan and Wang, Jiaqi and others},
  booktitle={Proceedings of the 32nd ACM International Conference on Multimedia},
  pages={11198--11201},
  year={2024}
}

@article{qwen2_5_vl,
  title={Qwen2. 5-vl technical report},
  author={Bai, Shuai and Chen, Keqin and Liu, Xuejing and Wang, Jialin and Ge, Wenbin and Song, Sibo and Dang, Kai and Wang, Peng and Wang, Shijie and Tang, Jun and others},
  journal={arXiv preprint arXiv:2502.13923},
  year={2025}
}

@article{huang2025vision,
  title={Vision-r1: Incentivizing reasoning capability in multimodal large language models},
  author={Huang, Wenxuan and Jia, Bohan and Zhai, Zijie and Cao, Shaosheng and Ye, Zheyu and Zhao, Fei and Xu, Zhe and Hu, Yao and Lin, Shaohui},
  journal={arXiv preprint arXiv:2503.06749},
  year={2025}
}

@misc{chen2025vinci,
  title={Vinci. R1-v: Reinforcing super generalization ability in vision-language models with less than \$3},
  author={Chen, Liang and Li, Lei and Zhao, Haozhe and Song, Yifan},
  journal={arXiv preprint arXiv:2308.15363},
  year={2025}
}

@article{zhang2025r1,
  title={R1-vl: Learning to reason with multimodal large language models via step-wise group relative policy optimization},
  author={Zhang, Jingyi and Huang, Jiaxing and Yao, Huanjin and Liu, Shunyu and Zhang, Xikun and Lu, Shijian and Tao, Dacheng},
  journal={arXiv preprint arXiv:2503.12937},
  year={2025}
}

@article{tan2025reason,
  title={Reason-rft: Reinforcement fine-tuning for visual reasoning},
  author={Tan, Huajie and Ji, Yuheng and Hao, Xiaoshuai and Lin, Minglan and Wang, Pengwei and Wang, Zhongyuan and Zhang, Shanghang},
  journal={arXiv preprint arXiv:2503.20752},
  year={2025}
}

@article{chen2025sft,
  title={Sft or rl? an early investigation into training r1-like reasoning large vision-language models},
  author={Chen, Hardy and Tu, Haoqin and Wang, Fali and Liu, Hui and Tang, Xianfeng and Du, Xinya and Zhou, Yuyin and Xie, Cihang},
  journal={arXiv preprint arXiv:2504.11468},
  year={2025}
}

@article{wang2025vl,
  title={Vl-rethinker: Incentivizing self-reflection of vision-language models with reinforcement learning},
  author={Wang, Haozhe and Qu, Chao and Huang, Zuming and Chu, Wei and Lin, Fangzhen and Chen, Wenhu},
  journal={arXiv preprint arXiv:2504.08837},
  year={2025}
}

@article{wang2025perception,
  title={Perception-aware policy optimization for multimodal reasoning},
  author={Wang, Zhenhailong and Guo, Xuehang and Stoica, Sofia and Xu, Haiyang and Wang, Hongru and Ha, Hyeonjeong and Chen, Xiusi and Chen, Yangyi and Yan, Ming and Huang, Fei and others},
  journal={arXiv preprint arXiv:2507.06448},
  year={2025}
}

@article{liu2025more,
  title={More Thinking, Less Seeing? Assessing Amplified Hallucination in Multimodal Reasoning Models},
  author={Liu, Chengzhi and Xu, Zhongxing and Wei, Qingyue and Wu, Juncheng and Zou, James and Wang, Xin Eric and Zhou, Yuyin and Liu, Sheng},
  journal={arXiv preprint arXiv:2505.21523},
  year={2025}
}

@article{yao2025reasoning,
  title={Are Reasoning Models More Prone to Hallucination?},
  author={Yao, Zijun and Liu, Yantao and Chen, Yanxu and Chen, Jianhui and Fang, Junfeng and Hou, Lei and Li, Juanzi and Chua, Tat-Seng},
  journal={arXiv preprint arXiv:2505.23646},
  year={2025}
}

@article{huang2025boosting,
  title={Boosting mllm reasoning with text-debiased hint-grpo},
  author={Huang, Qihan and Dai, Weilong and Liu, Jinlong and He, Wanggui and Jiang, Hao and Song, Mingli and Chen, Jingyuan and Yao, Chang and Song, Jie},
  journal={arXiv preprint arXiv:2503.23905},
  year={2025}
}

@article{shao2024deepseekmath,
  title={Deepseekmath: Pushing the limits of mathematical reasoning in open language models},
  author={Shao, Zhihong and Wang, Peiyi and Zhu, Qihao and Xu, Runxin and Song, Junxiao and Bi, Xiao and Zhang, Haowei and Zhang, Mingchuan and Li, YK and Wu, Yang and others},
  journal={arXiv preprint arXiv:2402.03300},
  year={2024}
}

@article{guo2025deepseek,
  title={Deepseek-r1: Incentivizing reasoning capability in llms via reinforcement learning},
  author={Guo, Daya and Yang, Dejian and Zhang, Haowei and Song, Junxiao and Zhang, Ruoyu and Xu, Runxin and Zhu, Qihao and Ma, Shirong and Wang, Peiyi and Bi, Xiao and others},
  journal={arXiv preprint arXiv:2501.12948},
  year={2025}
}

@inproceedings{thawakar2025llamav,
  title={Llamav-o1: Rethinking step-by-step visual reasoning in llms},
  author={Thawakar, Omkar and Dissanayake, Dinura and More, Ketan Pravin and Thawkar, Ritesh and Heakl, Ahmed and Ahsan, Noor and Li, Yuhao and Zumri, Ilmuz Zaman Mohammed and Lahoud, Jean and Anwer, Rao Muhammad and others},
  booktitle={Findings of the Association for Computational Linguistics: ACL 2025},
  pages={24290--24315},
  year={2025}
}

@article{xiao2025advancing,
  title={Advancing Multimodal Reasoning Capabilities of Multimodal Large Language Models via Visual Perception Reward},
  author={Xiao, Tong and Xu, Xin and Huang, Zhenya and Gao, Hongyu and Liu, Quan and Liu, Qi and Chen, Enhong},
  journal={arXiv preprint arXiv:2506.07218},
  year={2025}
}

@article{xia2025visionary,
  title={Visionary-r1: Mitigating shortcuts in visual reasoning with reinforcement learning},
  author={Xia, Jiaer and Zang, Yuhang and Gao, Peng and Li, Yixuan and Zhou, Kaiyang},
  journal={arXiv preprint arXiv:2505.14677},
  year={2025}
}

@article{zhang2025thyme,
  title={Thyme: Think beyond images},
  author={Zhang, Yi-Fan and Lu, Xingyu and Yin, Shukang and Fu, Chaoyou and Chen, Wei and Hu, Xiao and Wen, Bin and Jiang, Kaiyu and Liu, Changyi and Zhang, Tianke and others},
  journal={arXiv preprint arXiv:2508.11630},
  year={2025}
}

@article{chen2025mint,
  title={MINT-CoT: Enabling Interleaved Visual Tokens in Mathematical Chain-of-Thought Reasoning},
  author={Chen, Xinyan and Zhang, Renrui and Jiang, Dongzhi and Zhou, Aojun and Yan, Shilin and Lin, Weifeng and Li, Hongsheng},
  journal={arXiv preprint arXiv:2506.05331},
  year={2025}
}

@article{yang2025look,
  title={Look-back: Implicit visual re-focusing in mllm reasoning},
  author={Yang, Shuo and Niu, Yuwei and Liu, Yuyang and Ye, Yang and Lin, Bin and Yuan, Li},
  journal={arXiv preprint arXiv:2507.03019},
  year={2025}
}

@article{zhang2025deepsketcher,
  title={DeepSketcher: Internalizing Visual Manipulation for Multimodal Reasoning},
  author={Zhang, Chi and Qiu, Haibo and Zhang, Qiming and Zeng, Zhixiong and Ma, Lin and Zhang, Jing},
  journal={arXiv preprint arXiv:2509.25866},
  year={2025}
}

@article{zheng2025deepeyes,
  title={DeepEyes: Incentivizing" Thinking with Images" via Reinforcement Learning},
  author={Zheng, Ziwei and Yang, Michael and Hong, Jack and Zhao, Chenxiao and Xu, Guohai and Yang, Le and Shen, Chao and Yu, Xing},
  journal={arXiv preprint arXiv:2505.14362},
  year={2025}
}

@article{yu2025dapo,
  title={Dapo: An open-source llm reinforcement learning system at scale},
  author={Yu, Qiying and Zhang, Zheng and Zhu, Ruofei and Yuan, Yufeng and Zuo, Xiaochen and Yue, Yu and Dai, Weinan and Fan, Tiantian and Liu, Gaohong and Liu, Lingjun and others},
  journal={arXiv preprint arXiv:2503.14476},
  year={2025}
}

@article{li2025self,
  title={Self-rewarding vision-language model via reasoning decomposition},
  author={Li, Zongxia and Yu, Wenhao and Huang, Chengsong and Liu, Rui and Liang, Zhenwen and Liu, Fuxiao and Che, Jingxi and Yu, Dian and Boyd-Graber, Jordan and Mi, Haitao and others},
  journal={arXiv preprint arXiv:2508.19652},
  year={2025}
}

@article{lu2023mathvista,
  title={Mathvista: Evaluating mathematical reasoning of foundation models in visual contexts},
  author={Lu, Pan and Bansal, Hritik and Xia, Tony and Liu, Jiacheng and Li, Chunyuan and Hajishirzi, Hannaneh and Cheng, Hao and Chang, Kai-Wei and Galley, Michel and Gao, Jianfeng},
  journal={arXiv preprint arXiv:2310.02255},
  year={2023}
}

@article{wang2024measuring,
  title={Measuring multimodal mathematical reasoning with math-vision dataset},
  author={Wang, Ke and Pan, Junting and Shi, Weikang and Lu, Zimu and Ren, Houxing and Zhou, Aojun and Zhan, Mingjie and Li, Hongsheng},
  journal={Advances in Neural Information Processing Systems},
  volume={37},
  pages={95095--95169},
  year={2024}
}

@inproceedings{zhang2024mathverse,
  title={Mathverse: Does your multi-modal llm truly see the diagrams in visual math problems?},
  author={Zhang, Renrui and Jiang, Dongzhi and Zhang, Yichi and Lin, Haokun and Guo, Ziyu and Qiu, Pengshuo and Zhou, Aojun and Lu, Pan and Chang, Kai-Wei and Qiao, Yu and others},
  booktitle={European Conference on Computer Vision},
  pages={169--186},
  year={2024},
  organization={Springer}
}

@article{zou2024dynamath,
  title={Dynamath: A dynamic visual benchmark for evaluating mathematical reasoning robustness of vision language models},
  author={Zou, Chengke and Guo, Xingang and Yang, Rui and Zhang, Junyu and Hu, Bin and Zhang, Huan},
  journal={arXiv preprint arXiv:2411.00836},
  year={2024}
}

@inproceedings{qiao2025we,
  title={We-math: Does your large multimodal model achieve human-like mathematical reasoning?},
  author={Qiao, Runqi and Tan, Qiuna and Dong, Guanting and MinhuiWu, MinhuiWu and Sun, Chong and Song, Xiaoshuai and Wang, Jiapeng and Gongque, Zhuoma and Lei, Shanglin and Zhang, Yifan and others},
  booktitle={Proceedings of the 63rd Annual Meeting of the Association for Computational Linguistics (Volume 1: Long Papers)},
  pages={20023--20070},
  year={2025}
}

@article{xiao2024logicvista,
  title={Logicvista: Multimodal llm logical reasoning benchmark in visual contexts},
  author={Xiao, Yijia and Sun, Edward and Liu, Tianyu and Wang, Wei},
  journal={arXiv preprint arXiv:2407.04973},
  year={2024}
}

@misc{2023opencompass,
    title={OpenCompass: A Universal Evaluation Platform for Foundation Models},
    author={OpenCompass Contributors},
    howpublished = {\url{https://github.com/open-compass/opencompass}},
    year={2023}
}

@misc{zheng2025easyr1,
  title        = {EasyR1: An Efficient, Scalable, Multi-Modality RL Training Framework},
  author       = {Yaowei Zheng and Junting Lu and Shenzhi Wang and Zhangchi Feng and Dongdong Kuang and Yuwen Xiong},
  howpublished = {\url{https://github.com/hiyouga/EasyR1}},
  year         = {2025}
}

@article{sheng2024hybridflow,
  title   = {HybridFlow: A Flexible and Efficient RLHF Framework},
  author  = {Guangming Sheng and Chi Zhang and Zilingfeng Ye and Xibin Wu and Wang Zhang and Ru Zhang and Yanghua Peng and Haibin Lin and Chuan Wu},
  year    = {2024},
  journal = {arXiv preprint arXiv: 2409.19256}
}

@article{loshchilov2017decoupled,
  title={Decoupled weight decay regularization},
  author={Loshchilov, Ilya and Hutter, Frank},
  journal={arXiv preprint arXiv:1711.05101},
  year={2017}
}

@article{meng2025mm,
  title={Mm-eureka: Exploring visual aha moment with rule-based large-scale reinforcement learning},
  author={Meng, Fanqing and Du, Lingxiao and Liu, Zongkai and Zhou, Zhixiang and Lu, Quanfeng and Fu, Daocheng and Shi, Botian and Wang, Wenhai and He, Junjun and Zhang, Kaipeng and others},
  journal={CoRR},
  year={2025}
}

@article{lu2021inter,
  title={Inter-gps: Interpretable geometry problem solving with formal language and symbolic reasoning},
  author={Lu, Pan and Gong, Ran and Jiang, Shibiao and Qiu, Liang and Huang, Siyuan and Liang, Xiaodan and Zhu, Song-Chun},
  journal={arXiv preprint arXiv:2105.04165},
  year={2021}
}

@article{yao2024mulberry,
  title={Mulberry: Empowering mllm with o1-like reasoning and reflection via collective monte carlo tree search},
  author={Yao, Huanjin and Huang, Jiaxing and Wu, Wenhao and Zhang, Jingyi and Wang, Yibo and Liu, Shunyu and Wang, Yingjie and Song, Yuxin and Feng, Haocheng and Shen, Li and others},
  journal={arXiv preprint arXiv:2412.18319},
  year={2024}
}

@article{deng2025openvlthinker,
  title={Openvlthinker: An early exploration to complex vision-language reasoning via iterative self-improvement},
  author={Deng, Yihe and Bansal, Hritik and Yin, Fan and Peng, Nanyun and Wang, Wei and Chang, Kai-Wei},
  journal={arXiv preprint arXiv:2503.17352},
  year={2025}
}

@article{shen2025vlm,
  title={Vlm-r1: A stable and generalizable r1-style large vision-language model},
  author={Shen, Haozhan and Liu, Peng and Li, Jingcheng and Fang, Chunxin and Ma, Yibo and Liao, Jiajia and Shen, Qiaoli and Zhang, Zilun and Zhao, Kangjia and Zhang, Qianqian and others},
  journal={arXiv preprint arXiv:2504.07615},
  year={2025}
}

@article{yang2025wethink,
  title={WeThink: Toward General-purpose Vision-Language Reasoning via Reinforcement Learning},
  author={Yang, Jie and Ma, Feipeng and Wang, Zitian and Yin, Dacheng and Rong, Kang and Rao, Fengyun and Zhang, Ruimao},
  journal={arXiv preprint arXiv:2506.07905},
  year={2025}
}

@misc{openai2024gpt41,
  title        = {GPT-4.1},
  author       = {{OpenAI}},
  howpublished = {\url{https://openai.com/index/gpt-4-1/}},
year={2025},
}

@misc{XAI2024grok15v,
  title        = {grok-1.5v},
  author       = {{XAI}},
  howpublished = {\url{https://x.ai/news/grok-1.5v}},
year={2025},
}

@inproceedings{yue2024mmmu,
  title={Mmmu: A massive multi-discipline multimodal understanding and reasoning benchmark for expert agi},
  author={Yue, Xiang and Ni, Yuansheng and Zhang, Kai and Zheng, Tianyu and Liu, Ruoqi and Zhang, Ge and Stevens, Samuel and Jiang, Dongfu and Ren, Weiming and Sun, Yuxuan and others},
  booktitle={Proceedings of the IEEE/CVF Conference on Computer Vision and Pattern Recognition},
  pages={9556--9567},
  year={2024}
}

@inproceedings{yue2025mmmu,
  title={Mmmu-pro: A more robust multi-discipline multimodal understanding benchmark},
  author={Yue, Xiang and Zheng, Tianyu and Ni, Yuansheng and Wang, Yubo and Zhang, Kai and Tong, Shengbang and Sun, Yuxuan and Yu, Botao and Zhang, Ge and Sun, Huan and others},
  booktitle={Proceedings of the 63rd Annual Meeting of the Association for Computational Linguistics (Volume 1: Long Papers)},
  pages={15134--15186},
  year={2025}
}

@inproceedings{mathew2022infographicvqa,
  title={Infographicvqa},
  author={Mathew, Minesh and Bagal, Viraj and Tito, Rub{\`e}n and Karatzas, Dimosthenis and Valveny, Ernest and Jawahar, CV},
  booktitle={Proceedings of the IEEE/CVF Winter Conference on Applications of Computer Vision},
  pages={1697--1706},
  year={2022}
}

@inproceedings{paiss2023teaching,
  title={Teaching clip to count to ten},
  author={Paiss, Roni and Ephrat, Ariel and Tov, Omer and Zada, Shiran and Mosseri, Inbar and Irani, Michal and Dekel, Tali},
  booktitle={Proceedings of the IEEE/CVF International Conference on Computer Vision},
  pages={3170--3180},
  year={2023}
}

@inproceedings{kembhavi2016diagram,
  title={A diagram is worth a dozen images},
  author={Kembhavi, Aniruddha and Salvato, Mike and Kolve, Eric and Seo, Minjoon and Hajishirzi, Hannaneh and Farhadi, Ali},
  booktitle={European conference on computer vision},
  pages={235--251},
  year={2016},
  organization={Springer}
}

@inproceedings{fu2024blink,
  title={Blink: Multimodal large language models can see but not perceive},
  author={Fu, Xingyu and Hu, Yushi and Li, Bangzheng and Feng, Yu and Wang, Haoyu and Lin, Xudong and Roth, Dan and Smith, Noah A and Ma, Wei-Chiu and Krishna, Ranjay},
  booktitle={European Conference on Computer Vision},
  pages={148--166},
  year={2024},
  organization={Springer}
}

@article{team2025qwen3,
  title={Qwen3-vl: Sharper vision, deeper thought, broader action},
  author={Team, Qwen},
  journal={Qwen Blog. Accessed},
  pages={10--04},
  year={2025}
}
